\documentclass[11pt,a4paper]{article}
\usepackage{times,latexsym}
\usepackage{url}
\usepackage[T1]{fontenc}
\usepackage[whole]{bxcjkjatype}

\usepackage[acceptedWithA]{tacl2021v1}

\usepackage{xspace,mfirstuc,tabulary}

\newcommand{\ex}[1]{{\sf #1}}

\newif\iftaclinstructions
\taclinstructionsfalse 
\iftaclinstructions
\renewcommand{\confidential}{}
\renewcommand{\anonsubtext}{(No author info supplied here, for consistency with
TACL-submission anonymization requirements)}
\newcommand{\instr}
\fi

\iftaclpubformat 

\else

\fi

\usepackage{microtype}
\usepackage{setspace}
\usepackage{color}
\usepackage{amssymb}
\usepackage{gb4ea}
\usepackage{multirow}
\usepackage{tikz,subfigure}
\usetikzlibrary{automata,positioning,arrows,shapes,backgrounds,trees}
\usepackage{tikz-qtree}
\usepackage{tikz-qtree-compat}
\usepackage{xcolor}
\definecolor{lightgray}{rgb}{0.8, 0.8, 0.8}
\pgfdeclarelayer{background}
\pgfdeclarelayer{foreground}
\pgfsetlayers{background,main,foreground}

\newcommand{\lowacc}[1]{\textcolor{red}{#1}}

\newcommand{\lentail}{\textit{entailment}}
\newcommand{\lneutral}{\textit{neutral}}
\newcommand{\lcontra}{\textit{contradiction}}
\newcommand{\bertja}{jaBERT}
\newcommand{\bertm}{mBERT}
\newcommand{\subword}{\textsc{(subword)}}
\newcommand{\charact}{\textsc{(char)}}
\newcommand{\wholema}{\textsc{(whole)}}
\newcommand{\spair}{(\textit{A}, \textit{B})}
\newcommand{\premise}{\textit{A}}
\newcommand{\hypothesis}{\textit{B}}
\newcommand{\SFord}{\mathsf{ord}}
\newcommand{\SFcase}{\mathsf{case}}
\newcommand{\SFdel}{\mathsf{del}}
\newcommand{\premord}{\textit{A}$_{\SFord}$}
\newcommand{\premcase}{\textit{A}$_{\SFcase}$}
\newcommand{\premdel}{\textit{A}$_{\SFdel}$}
\newcommand{\scrampair}{(\textit{A}$_{\SFord}$, \textit{B})}
\newcommand{\swappair}{(\textit{A}$_{\SFcase}$, \textit{B})}
\newcommand{\delpair}{(\textit{A}$_{\SFdel}$, \textit{B})}
\newcommand{\smtg}[1]{\ensuremath{\mathtt{#1}}} 
\newcommand{\semtag}[2][\footnotesize]{\tikz[baseline=0ex]\node[
	draw=gray,
	fill=green!20,
	inner sep=1.2pt, 
	anchor=text, 
	rectangle,
	rounded corners=.4mm]
{#1\smtg{#2}};} 
\newcommand{\tag}[2]{[#1$_{\semtag{#2}}$]}

\newcommand{\NPone}[1]{\textcolor{blue}{#1}}
\newcommand{\NPtwo}[1]{\textcolor{red}{#1}}
\newcommand{\TVo}[1]{\textcolor{orange}{#1}}
\newcommand{\Nom}[1]{\textcolor{green!60!black}{#1}}
\newcommand{\Acc}[1]{\textcolor{gray}{#1}}

\newcommand{\diffex}[5]{
\multirow{4}{*}{#1} &
$A$: #2 \\
&
\hspace{3ex} #3 \\
&
$B$: #4 \\
&
\hspace{3ex} #5 \\
}

\title{Compositional Evaluation on Japanese Textual Entailment and Similarity}

\author{
  \parbox{\linewidth}{\centering
   Hitomi Yanaka$^1$ and
   Koji Mineshima$^2$
  }
  \\
   $^1$\mbox{\rm The University of Tokyo,}
   $^2$\mbox{\rm Keio University}
  \\
  \parbox{\linewidth}{\centering
   \texttt{hyanaka@is.s.u-tokyo.ac.jp},
   \texttt{minesima@abelard.flet.keio.ac.jp}
   }
}

\date{}

\begin{document}
\maketitle
\begin{abstract}
Natural Language Inference (NLI) and Semantic Textual Similarity (STS) are widely used benchmark tasks for compositional evaluation of pre-trained language models.
Despite growing interest in linguistic universals, most NLI/STS studies have focused almost exclusively on English.
In particular, there are no available multilingual NLI/STS datasets in Japanese, which is typologically different from English and can shed light on the currently controversial behavior of language models in matters such as sensitivity to word order and case particles.
Against this background, we introduce JSICK, a Japanese NLI/STS dataset that was manually translated from the English dataset SICK.
We also present a stress-test dataset for compositional inference, created by transforming syntactic structures of sentences in JSICK to investigate whether language models are sensitive to word order and case particles. 
We conduct baseline experiments on different pre-trained language models and compare the performance of multilingual models when applied to Japanese and other languages.
The results of the stress-test experiments suggest that the current pre-trained language models are insensitive to word order and case marking.
\end{abstract}

\section{Introduction}
\label{sec:intro}
Natural Language Inference (NLI)~\cite{dagan2006,bowman-etal-2015-large} and
Semantic Textual Similarity (STS)~\cite{agirre-etal-2016-semeval} tasks are well-positioned to serve as a basic benchmark for natural language understanding.
With the recent progress of deep neural networks including pre-trained language models such as BERT~\cite{devlin-etal-2019-bert},
the development of benchmark datasets has centered on large crowdsourced English datasets, such as SNLI~\cite{bowman-etal-2015-large} and MultiNLI~\cite{williams-etal-2018-broad}.
Since there has been an increasing need for benchmark datasets in linguistic universals~\cite{linzen-2020-accelerate}, general language understanding frameworks including NLI and STS for languages other than English have been provided~\cite{conneau-etal-2018-xnli,liang-etal-2020-xglue,le-etal-2020-flaubert,shavrina-etal-2020-russiansuperglue,xu-etal-2020-clue,seelawi-etal-2021-alue,park2021klue}.

Another recent line of work has investigated whether models are sensitive to shuffled word order, but the conclusions are controversial~\cite{ravfogel-etal-2019-studying,sinha2021masked,sinha2021unnat,DBLP:conf/aaai/GuptaKS21,pham-etal-2021-order,white-cotterell-2021-examining}.
One characteristic of human-like language understanding is that humans can understand sentences according to their word meanings and syntactic structures, then recognize their semantic relationships~\cite{frege1963compound,katz1963structure,Montague:1973,janssen1997compositionality}.
Since previous work has demonstrated the usefulness of analyzing the generalization ability of models in challenging NLI in English~\cite{naik-etal-2018-stress,glockner-etal-2018-breaking,mccoy-etal-2019-right,rozen-etal-2019-diversify,goodwin-etal-2020-probing,yanaka-etal-2021-exploring}, we should continue this line of research in other languages.

Against this background, we provide a Japanese NLI/STS dataset to analyze language models in compositional inference across languages.
Our motivations for focusing on Japanese are two-fold.
First, Japanese is a high-resource language that has typologically different characteristics from English~\cite{joshi-etal-2020-state}, yet it has not been included in previous cross-lingual~\cite{real2018sickBR,hu-etal-2020-ocnli,ham-etal-2020-kornli,wijnholds2021sicknl} or multilingual~\cite{conneau-etal-2018-xnli} NLI datasets.
This raises the question of whether models perform inference differently in Japanese and other languages.
Second, Japanese has case markers and free word-order~\cite{hinds1986,shibatani1990languages},
phenomena that pose interesting challenges for multilingual NLI.
While shuffling data usually changes its meaning, the meaning of a Japanese sentence can be preserved even when the order of noun phrase (NP) arguments is swapped.
By analyzing model behavior with scrambling phenomena that preserve case relations in a sentence and particle-swapping phenomena that change case relations, we can analyze whether the model can distinguish transformations that change sentence meanings and perform compositional inferences.

This paper has three contributions.
First,
we provide JSICK as a compositional Japanese NLI/STS dataset by manually translating the English SICK dataset~\cite{marelli-etal-2014-sick}.
Compared with recent crowdsourced NLI datasets,
SICK facilitates identification of which compositional linguistic phenomena are key to a given inference.
Such a controlled structure is suited to transforming sentences for further analyses of model behavior.
In addition, SICK has been translated into non-English languages~\cite{real2018sickBR,wijnholds2021sicknl}, allowing cross-language comparisons
on a sizeable parallel NLI corpus.

Second, we create a stress-test dataset for JSICK to investigate whether language models capture word order and case particles in Japanese.
We created the stress-test dataset by
transforming syntactic structures of JSICK sentence pairs, where we analyze whether models consider word order and case particles when predicting entailment labels and similarity scores.

Third, for the baseline evaluation of pre-trained language models, we compare performance between different pre-trained language models on JSICK.
We also compare the performance of multilingual pre-trained language models on SICK datasets of different languages, including JSICK.
We also provide an in-depth analysis of sensitivity to word order and case particles
based on the JSICK stress-test dataset.
The analysis results suggest that both Japanese and multilingual models are surprisingly inattentive to word order and case marking.
Our dataset will be publicly available at \url{https://github.com/verypluming/JSICK}.

\section{Related Work}
\label{sec:related}
Standard NLI benchmarks have been mainly developed for English.
Recently, large crowdsourced NLI datasets derived from image captions, such as 
SNLI~\cite{bowman-etal-2015-large},
and
those targeting multi-genre sentences, like MultiNLI~\cite{williams-etal-2018-broad}, have been widely used to evaluate neural models.
For linguistics-oriented datasets, FraCaS~\cite{cooper1994fracas} is a manually collected NLI test set involving linguistic phenomena studied in formal semantics, and
SICK~\cite{marelli-etal-2014-sick} is a larger and more naturalistic NLI/STS dataset made from captions focusing on compositional inference.
Unlike SNLI and MultiNLI, SICK was designed by linguistic experts so as to not require dealing with aspects beyond the scope of compositional inference (e.g., world knowledge, named entities, and multiword expressions) but to cover a variety of combinations of lexical, syntactic, and semantic phenomena.
The SICK dataset thus allows systematic assessments of the reasoning ability of models on compositional inference.
For STS, the SemEval 2012--2017~\cite{agirre-etal-2016-semeval,cer-etal-2017-semeval} competitions provided English, Arabic, and Spanish STS datasets including SICK.

With the development of multilingual pre-trained language models,
general language understanding frameworks for languages other than English have been created~\cite{liang-etal-2020-xglue,le-etal-2020-flaubert,shavrina-etal-2020-russiansuperglue,xu-etal-2020-clue,seelawi-etal-2021-alue,park2021klue}, and NLI datasets have been multilingualized.
\citet{conneau-etal-2018-xnli} provided a cross-lingual NLI (XNLI) corpus by translating MultiNLI into
15 languages, including languages with few language resources such as Swahili and Urdu and languages with flexible word order such as Russian and German.
\citet{ham-etal-2020-kornli} translated MultiNLI into Korean to create KorNLI.
However, Japanese is not included in these datasets.
In addition, since sentences in MultiNLI are usually longer than those in SICK and contain multiword expressions beyond the scope of compositional inference, it is unrealistic to carefully transform syntactic structures of sentence pairs in XNLI to create a stress-test dataset like ours.

As examples of other non-English datasets, 
OCNLI~\cite{hu-etal-2020-ocnli} is a Chinese NLI dataset built from original multi-genre resources.
FarsTail~\cite{amirkhani2020farstail} is a Persian NLI dataset containing sentences from university exams.
There have also been attempts to translate the SICK dataset into Portuguese~\cite{real2018sickBR} and Dutch~\cite{wijnholds2021sicknl}, so our Japanese SICK dataset will contribute to a multilingual SICK dataset that will allow controlled, cross-lingual analyses of the compositional abilities of language models.
 
Regarding Japanese NLI datasets, a
Japanese SNLI dataset~\cite{jsnli} was constructed by using machine translation to translate the English SNLI dataset into Japanese and automatically filtering out unnatural sentences, but methods that employ machine translation are still problematic in that they can produce unnatural sentences.
The Japanese Realistic Textual Entailment Corpus~\cite{hayashibe-2020-japanese} is a crowdsourced dataset containing Japanese hotel reviews.
However, linguistic phenomena in these Japanese datasets demonstrate limited diversity because the sentences they contain are restricted to simple structures.
\citet{10.1007/978-3-319-50953-2_5} provided JSeM,
a manually-curated test set including a Japanese version of FraCaS to diagnose inference systems from a formal semantics perspective.
We produced an
NLI dataset by asking experts to translate the SICK dataset into Japanese, thus maintaining both sentence naturalness and compositions of linguistic phenomena.

While recent works~\cite{sinha2021masked,sinha2021unnat,DBLP:conf/aaai/GuptaKS21,pham-etal-2021-order,hessel-schofield-2021-effective} have shown that pre-trained language models are insensitive to word order on permuted English datasets in the standard natural language understanding benchmark GLUE~\cite{wang2018glue} including NLI, other works have analyzed sentence perplexity with varying word orders and shown controversial results regarding inductive biases for word order in different languages~\cite{ravfogel-etal-2019-studying,white-cotterell-2021-examining}.
For six languages, including Japanese, \citet{yang-etal-2019-paws} evaluated whether multilingual BERT captures word order in the translated PAWS dataset~\cite{zhang-etal-2019-paws}, involving adversarial paraphrase identification pairs whose sentences share words but differ in word order.
Experiments showed that BERT performance for Japanese is consistently
worse than that for Indo-European languages.
Our study deepens insight into the causes of performance differences by stress-test evaluation of NLI and STS tasks through careful manipulation of case markers, which are more challenging tasks than are two-class paraphrase identification tasks.
\citet{kuribayashi-etal-2021-lower} reexamined the general hypothesis that language models with lower perplexity are more human-like in Japanese than in English, and the results have shown the necessity of evaluating models across languages.
Analyzing the model behavior, rather than perplexity, in transformed Japanese inference should provide new insights into the model's sensitivity to word order.

\section{JSICK Dataset Creation}
\label{sec:jsick}

\begin{table*}[h!t]
    \centering
        \scalebox{0.65}{
    \begin{tabular}{lp{36em}lll}\hline
    \textbf{ID}&\textbf{premise--hypothesis pair} \spair{}&\textbf{Linguistic phenomena}&\textbf{Entailment}&\textbf{Similarity} \\ \hline
     5862&\begin{tabular}{lp{32em}}\premise{}:& \tag{二人}{NUM}の女性が群衆の前でダンスをし\tag{ながら}{CONJ}歌っている\\
     &\tag{Two}{NUM} women are dancing \tag{and}{CONJ} singing in front of a crowd\\
     \hypothesis{}:& \tag{二人}{NUM}の女性が\tag{多く}{QUANT}の人の前でダンスをし\tag{ながら}{CONJ}歌っている\\
     &\tag{Two}{NUM} women are dancing \tag{and}{CONJ} singing in front of \tag{many}{QUANT} people
     \end{tabular}
     &\texttt{QUANT}/\texttt{NUM}/\texttt{CONJ}&\lentail{}&4.7 \\ \hline
     352&\begin{tabular}{lp{32em}}\premise{}:& \tag{別}{NUM}の犬を追いかけている\tag{か}{DISJ}、または口に棒きれをくわえている犬は\tag{一}{NUM}匹もい\tag{ない}{NEG}\\
     &There is \tag{no}{NEG} dog chasing another \tag{or}{DISJ} holding a stick in its mouth\\
     \hypothesis{}:& 犬が\tag{別}{NUM}の犬を追いかけていて、口に棒きれをくわえている\\
     &A dog is chasing \tag{another}{NUM} and is holding a stick in its mouth
     \end{tabular}
     &\texttt{NUM}/\texttt{DISJ}/\texttt{NEG}&\lcontra{}&3.9 \\ \hline
     765&\begin{tabular}{lp{32em}}\premise{}:& その子供は幸せ\tag{そう}{MODAL}に雪の中で滑っている\\
     &The kid is \tag{happily}{MODAL} sliding in the snow\\
     \hypothesis{}:&雪で覆わ\tag{れ}{PAS}た丘の上にいる男の子が赤いジャケット\tag{と}{CONJ}黒い帽子を身に着け、ひざまづいて滑っている\\
     &A boy on a hill \tag{covered}{PAS} in snow is wearing a red jacket \tag{and}{CONJ} a black hat and is sliding on his knees
     \end{tabular}
     &\texttt{MODAL}/\texttt{CONJ}/\texttt{PAS}&\lneutral{}&2.5\\
    \hline
    \end{tabular}
    }
    \caption{Examples from the JSICK dataset. Each ID corresponds to the ID in the original SICK dataset.}
    \label{tab:example}
\end{table*}

\begin{table}[h!t]
    \centering
    \scalebox{0.80}{
    \begin{tabular}{l|llll}\hline
         \textbf{Phenomenon}&\textbf{Train}&\textbf{Dev}&\textbf{Test}&\textbf{Total}  \\ \hline
         Numeral (\texttt{NUM})&1,374&151&1,513&3,038\\
         Negation (\texttt{NEG})&1,096&107&1,140&2,343 \\
         Quantification (\texttt{QUANT})&698&81&744&1,523\\
         Passive (\texttt{PAS})&649&83&695&1,427 \\ 
         Anaphora (\texttt{ANA})&612&74&700&1,386 \\ 
         Conjunction (\texttt{CONJ})&558&69&640&1,267\\
         Disjunction (\texttt{DISJ})&364&53&428&845\\
         Modal (\texttt{MODAL})&62&4&69&135 \\ 
         Additive particle (\texttt{ADDP})&7&1&13&21 \\ 
         \hline
    \end{tabular}
    }
    \caption{Distribution of linguistic tags in JSICK.}
    \label{tab:semtag}
\end{table}

\subsection{Translation}
\label{ssec:trans}
The original SICK dataset uses 6,077 sentences to provide 9,927 sentence pairs \spair{}.
To create the JSICK dataset, we first asked an expert translator to translate the 6,077 English sentences in SICK into Japanese.
The translator did not see entailment labels, instead just translating a list of English sentences sorted alphabetically.
The translations were independently validated by English--Japanese bilinguals, and no examples were discarded.
To avoid changing sentence meaning during translation, we prepared translation guidelines and asked the translator to translate English sentences into natural Japanese while maintaining diversity in lexical, syntactic, and semantic phenomena such as hypernym--hyponym relations, active--passive alternations, and quantification in the original English sentences.
Note that sentences in the JSICK dataset contain some translations that are unnatural due to cultural factors, but reflecting culture in translation is beyond the scope of analyzing the compositional inference ability of models.

The guidelines explain how to translate linguistic phenomena, including indefinite and definite articles, singular and plural nouns, passive verbs, negation,
and quantification.
We also asked the translator to try to keep word orders as consistent as possible with the original sentences.
Some instructions from our guidelines are given in detail below.

\paragraph{Indefinite/definite articles}
The following describes our instructions regarding the distinction between indefinite and definite articles.
The distinction between indefinite and definite articles is an important phenomenon that affects interpretations of quantification~\cite{Hawkins1978,Heim1982}.
However, since Japanese does not have articles~\cite{hinds1986,shibatani1990languages}, it is not obvious how to translate indefinite articles as in (\ref{ex:flute}) and definite articles as in (\ref{ex:guitar}) into Japanese.
We therefore translated subject NPs as bare noun phrases, using the particle が (\textit{ga}) when translating the nominative case involving an indefinite article, and using the particle (topic marker) は (\textit{wa}) when translating the nominative case involving a definite article.
Since the majority of sentences in the SICK dataset are episodic, we can correctly translate English sentences into Japanese by the above rule.
\begin{exe}
\ex 
\gll 男性 \textbf{が} ギター を 弾いて い ない\\
man Nom guitar Acc playing is not\\
\glt `\textbf{A} man is not playing a guitar' 
\label{ex:flute}
\ex 
\gll 男性 \textbf{は} ギター を 弾いて い ない\\
man Topic guitar Acc playing is not\\
\glt `\textbf{The} man is not playing a guitar'
\label{ex:guitar}
\end{exe}

\paragraph{Singular and plural nouns}
In examples like (\ref{ex:men}), we can translate plural nouns by adding a plurality suffix such as たち (\textit{-tachi}).
However, Japanese does not have a general way to form plural words like the \textit{-s} suffix in English.
Thus, as in (\ref{ex:boil}), we prioritized sentence naturalness by not adding the plural suffix たち to the accusative case of words like エビ (\textit{shrimps}).
\begin{exe}
\ex 
\gll 男性たち が 木 を 切って いる\\
men Nom wood Acc cutting are\\
\glt `Men are cutting wood' 
\label{ex:men}
\ex 
\gll 女性 が エビ を ゆでて いる\\
woman Nom shrimps Acc boiling is\\
\glt `A woman is boiling shrimps'
\label{ex:boil}
\end{exe}

\begin{table}[h!t]
    \centering
    \scalebox{0.65}{
    \begin{tabular}{c|llll}\hline
         \textbf{Gold label}&\textbf{Train}&\textbf{Dev}&\textbf{Test}&\textbf{Total}  \\ \hline
         \multicolumn{5}{c}{JSICK-NLI}\\ \hline
         Entailment&969 (21.5)&122 (24.4)&1,088 (22.1)&2,179 (22.0)\\
         Contradiction&743 (16.5)&80 (16.0)&797 (16.2)&1,620 (16.3)\\
         Neutral&2,788 (62.0)&298 (59.6)&3,042 (61.7)&6,128 (61.7)\\ \hline
         \multicolumn{5}{c}{SICK-NLI}\\ \hline
         Entailment&1,299 (28.9)&144 (28.8)&1,414 (28.7)&2,857 (28.8)\\
         Contradiction&665 (14.8)&74 (14.8)&720 (14.6)&1,459 (14.7)\\
         Neutral&2,536 (56.4)&282 (56.4)&2,793 (56.7)&5,611 (56.5)\\ \hline
         \multicolumn{5}{c}{JSICK-STS}\\ \hline
         1-2&614 (13.6)&71 (14.2)&651 (13.2)&1,336 (13.4)\\
         2-3&1,164 (25.9)&111 (22.2)&1,248 (25.3)&2,523 (25.4)\\
         3-4&1,373 (30.5)&155 (31.0)&1,587 (32.2)&3,115 (31.4)\\
         4-5&1,349 (30.0)&163 (32.6)&1,441 (29.2)&2,955 (29.7)\\ \hline
         \multicolumn{5}{c}{SICK-STS}\\ \hline
         1-2&436 (9.7)&37 (7.4)&451 (9.2)& 924(9.3)\\
         2-3&635 (14.1)&69 (13.8)&674 (13.7)&1,378 (13.9)\\
         3-4&1,742 (38.7)&192 (38.4)&1,965 (39.9)&3,899 (39.3)\\
         4-5&1,687 (37.5)&202 (40.4)&1,837 (37.3)&3,726 (37.5)\\ \hline
         Total&4,500&500&4,927&9,927\\
         \hline
    \end{tabular}
    }
    \caption{Distribution of JSICK and SICK sentence pairs for each gold
entailment label and similarity score. Numbers in parentheses are percentages of the entire dataset.}
    \label{tab:jsick}
\end{table}

\begin{table}[h!t]
    \centering
    \scalebox{0.68}{
    \begin{tabular}{c|lll|l}\hline
         \textbf{Similarity}&\textbf{Entailment}&\textbf{Contradiction}&\textbf{Neutral}&\textbf{Total}  \\ \hline
         1-2&0 (0.0)&5 (0.3)&1,331 (21.7)&1,336\\
         2-3&6 (0.3)&160 (9.9)&2,357 (38.4)&2,523\\
         3-4&293 (13.4)&633 (39.1)&2,189 (35.7)&3,115\\
         4-5&1,880 (86.3)&822 (50.7)&253 (4.1)&2,955\\ \hline
         Total&2,179&1,620&6,128&\\ 
         \hline
    \end{tabular}
    }
    \caption{Distribution of JSICK sentence pairs across NLI and STS tasks.}
    \label{tab:rtests}
\end{table}

\subsection{Validation}
\label{ssec:label}
There are issues with the gold labels in the original SICK
dataset~\cite{bowman-etal-2015-large,kalouli-etal-2017-textual}.
In addition, translation from English to Japanese can change the appropriateness of the entailment label used in English (see Section~\ref{sec:dataset}).
Thus, instead of using the original gold labels,
we used the crowdsourcing platform Lancers\footnote{\url{https://www.lancers.jp/}} to re-annotate entailment labels and similarity scores for JSICK. 
Definitions of entailment labels (\lentail{}, \lcontra{}, and \lneutral{}) and similarity scores in a range of 1 (completely unrelated) to 5 (very related) for a pair \spair{} of sentences are the same as those for the original SICK.
In our instructions, we noted that sentences \textit{A} and \textit{B} describe the same situation or event to avoid any indeterminacy of event and entity coreference
that might cause inconsistencies in \lcontra\ labels~\cite{bowman-etal-2015-large}.

The annotators were six native Japanese speakers, randomly selected from the crowdsourcing platform.
The authors annotated the gold labels with ten examples in the JSICK trial set (500 examples) to provide ten test questions.
We asked the annotators to fully understand the guidelines to the point where they could assign the same labels as gold labels for all ten test questions.
We adopted annotations that were agreed upon by a majority vote as gold entailment labels and adopted the average of the annotation results as gold similarity scores.
For entailment labels, the authors also manually checked whether the majority judgement vote was semantically valid for each example.
Since recent works~\cite{10.1162/tacl_a_00293,gantt-etal-2020-natural} have demonstrated the importance of information for modeling disagreements
in NLI datasets, we will publicly release the raw annotations with the JSICK dataset.

The average annotation time was 1 min per pair, and Krippendorff's alpha for the entailment labels was 0.65.
There were 6,957 cases (70.1\%) in which three annotators assigned the same entailment labels, 2,922 cases (29.5\%) in which two annotators assigned the same entailment labels, and 48 cases (0.4\%) in which the labels of all three annotators assigned different labels. 
For cases where the labels of all three annotators assigned different labels, the labels were determined by the consensus of the authors.

\begin{table}[h!t]
    \centering
    \scalebox{0.77}{
    \begin{tabular}{l|lll}\hline
         \textbf{Phenomenon}&\textbf{JSICK}&\textbf{JSNLI}&\textbf{JRTEC}  \\ \hline
         Numeral&1,513 (30.7)&1,030 (26.3)&47 (1.2)\\
         Negation&1,140 (23.1)&66 (1.7)&291 (7.5)\\
         Quantification&744 (15.1)&298 (7.6)&185 (4.8)\\
         Passive&695 (14.1)&226 (5.8)&89 (2.3)\\ 
         Anaphora&700 (14.2)&487 (12.4)&72 (1.9)\\ 
         Conjunction&640 (13.0)&922 (23.5)&136 (3.5)\\
         Disjunction&428 (8.7)&168 (4.3)&65 (1.7)\\
         Modal&69 (1.4)&103 (2.6)&11 (0.3)\\ 
         Additive particle&13 (0.3)&6 (0.2)&39 (1.0)\\ \hline
         Test total&4,927&3,916&3,885\\ 
         \hline
    \end{tabular}
    }
    \caption{Comparison of linguistic tags between JSICK and previous Japanese NLI datasets. Numbers in parentheses are percentages of the entire test set.}
    \label{tab:jsicksnli}
\end{table}

\subsection{Linguistic tags}
\label{ssec:sem}
To analyze the ability of models to capture various linguistic phenomena, we annotated the JSICK dataset with linguistic phenomenon tags.
We provided a set of nine linguistic tags for linguistic phenomena: numerals, negation, quantification, passive voices, anaphora, conjunction, disjunction, modal, and additive particle.
We automatically annotated multiple tags with premise--hypothesis pairs, using Janome\footnote{\url{https://github.com/mocobeta/janome}} to process each premise and hypothesis sentence for morphological analysis and part-of-speech tagging.
If results of either of the morphological analyses included phrase patterns related to a linguistic tag, the premise--hypothesis pair is annotated with that tag.

Table~\ref{tab:example} shows examples of linguistic tagging in the JSICK dataset, and Table~\ref{tab:semtag} shows the distribution of linguistic tags.
Table~\ref{tab:jsicksnli} shows the results of comparing the percentage of linguistic tags in the JSICK test data and the two existing large Japanese NLI datasets mentioned in Section~\ref{sec:related}, Japanese SNLI (JSNLI) and the Japanese Realistic Textual Entailment Corpus (JRTEC).
Compared with previous datasets, JSICK contains more linguistic phenomena, including numerals, negation, quantification, passive voice, anaphora, and disjunction. 
This indicates that the distribution of linguistic phenomena in the JSICK dataset is well balanced.

\subsection{Dataset}
\label{sec:dataset}
Table~\ref{tab:jsick} shows that the distribution of JSICK dataset gold labels is almost the same as that for the English SICK dataset.
The distribution of JSICK sentence pairs across NLI and STS tasks (Table~\ref{tab:rtests}) also follows the same trend as in SICK; similarity scores for the \lentail{} and \lcontra{} cases tend to be in the range of 3 to 5, while \lneutral{} similarity scores are distributed.

\begin{table*}[h!t]
\centering
\scalebox{0.65}{
\begin{tabular}{l | l} \hline
\textbf{Type} 
& \textbf{Example (ID)} \\ \hline
\diffex{Singular-Plural}
{ある \ 人 \ が \ \textbf{キノコ} \ を \ ナイフ \ で \ 切って \ いる
(\textit{A person is cutting a mushroom with a knife})}
{a person Nom mushroom Acc knife with cutting is}
{ある \ 人 \ が \ \textbf{いくつかの} \ \textbf{キノコ} \ を \ 切って \ いる 
(\textit{A person is cutting some mushrooms})}
{a person Nom some mushroom Acc cutting is
\hfill{ID: JSICK-2590}} \hline
\diffex{Lexical Gap}
{ある \ \textbf{人} \ が \ ピンク色 \ の \ ロープ \ で \ 岩 \ を \ 登って \ いる
(\textit{A person is climbing a rock with a rope, which is pink})}
{a person Nom pink is rope with rock Acc climbing is}
{一人の \ \textbf{男性} \ が \ ロープ \ で \ その \ 崖 \ を \ 登って \ いる 
(\textit{One man is climbing the cliff with a rope})}
{one man Nom rope with the cliff Acc climbing is
\hfill{ID: JSICK-648}} \hline
\end{tabular}
}
\caption{Examples of linguistic factors that cause differences in entailment labels between English and Japanese.}
\label{tab:GrammarDiff}
\end{table*}

The most common cases where translation changed the entailment label from that in the original SICK dataset were those where the labels are changed to \lneutral.
There were 242 such examples, due to grammatical differences between English and Japanese.
Table~\ref{tab:GrammarDiff} shows some typical examples.
One major grammatical difference
that can change entailment labels
is the distinction between singular and plural NPs.
In English, the plural form \textit{mushrooms} explicitly indicates that there is more than one mushroom.
By contrast, there is no grammatical singular-plural marking in Japanese~\cite{nakanishi2004japanese}, so the bare noun キノコ (``mushroom'')
can be interpreted as either singular or plural.
This caused split entailment judgments among the annotators.
Other types of discrepancy are due to
various lexical gaps.
For instance, 
in Lexical Gap Example~B in Table~\ref{tab:GrammarDiff},
the English word \textit{man} can be applied to both men and women, while its natural counterpart (男性) in Japanese does not have such a generic meaning.
As a result, the entailment label for this example is \lneutral, rather than \lentail.

\begin{table*}[h!t]
    \centering
    \scalebox{0.77}{
    \begin{tabular}{lllcccc}\hline
    \textbf{Model}&\textbf{Train (setting)}&\textbf{Test}&\textbf{Prec}&\textbf{Rec}&\textbf{macro-F1}&\textbf{Acc}\\ \hline \hline
    \multicolumn{7}{c}{\textbf{Japanese models}} \\ \hline
    \multirow{6}{*}{jaRoBERTa-large}&JSICK&JSICK&90.1&87.3&88.6&90.3$\pm$0.04\\
    &JSICK (hypo-only)&JSICK&21.0&33.3&25.7&62.9$\pm$0.09\\
    &JSNLI&JSICK&61.8&69.3&54.9&53.8$\pm$0.09\\
    &&JSNLI&94.2&94.4&94.2&94.3$\pm$0.06\\
    &JSNLI+JSICK&JSICK&86.7&88.2&87.4&89.0$\pm$0.09\\ 
    &&JSNLI&93.6&93.7&93.6&93.7$\pm$0.09\\ \hline
    \multirow{6}{*}{jaRoBERTa-base}&JSICK&JSICK&84.9&87.8&86.2& 87.9$\pm$0.07\\
    &JSICK (hypo-only)&JSICK&21.0&33.3&25.7&62.9$\pm$0.08\\
    &JSNLI&JSICK&60.2&67.4&53.5&52.6$\pm$0.07\\
    &&JSNLI&93.0&93.0&92.9&93.0$\pm$0.06\\
    &JSNLI+JSICK&JSICK&83.0&90.1&85.5&87.9$\pm$0.05\\
    &&JSNLI&92.4&92.5&92.4&92.5$\pm$0.03\\ \hline
    \multirow{6}{*}{jaBERT-large}&JSICK&JSICK&86.5&84.8&85.6&87.9$\pm$0.03\\
    &JSICK (hypo-only)&JSICK&20.6&33.3&25.5&61.8$\pm$0.09\\
    &JSNLI&JSICK&57.8&64.9&52.5&52.2$\pm$0.05\\
    &&JSNLI&92.6&92.7&92.6&92.7$\pm$0.04\\
    &JSNLI+JSICK&JSICK&88.1&88.7&88.4&90.0$\pm$0.03\\ 
    &&JSNLI&93.4&93.5&93.4&93.5$\pm$0.02\\ \hline
    \multirow{6}{*}{jaBERT-base \wholema{}}&JSICK&JSICK&78.6&81.8&79.9&82.4$\pm$0.05\\
    &JSICK (hypo-only)&JSICK&48.8&42.8&44.0&58.9$\pm$0.07\\
    &JSNLI&JSICK&57.5&63.7&52.5&52.4$\pm$0.05\\
    &&JSNLI&94.1&94.2&94.1&94.2$\pm$0.03\\
    &JSNLI+JSICK&JSICK&84.4&87.0&85.6&87.5$\pm$0.03\\ 
    &&JSNLI&94.3&94.3&94.3&94.3$\pm$0.03\\ \hline
    \multirow{6}{*}{jaBERT-base \charact{}}&JSICK&JSICK&76.2&80.6&78.1&80.7$\pm$0.05\\
    &JSICK (hypo-only)&JSICK&47.8&46.6&47.0&54.9$\pm$0.08\\
    &JSNLI&JSICK&55.5&60.2&47.9&47.9$\pm$0.03\\
    &&JSNLI&90.7&90.8&90.7&90.8$\pm$0.03\\
    &JSNLI+JSICK&JSICK&83.8&85.3&84.4&86.3$\pm$0.04\\ 
    &&JSNLI&90.8&90.8&90.8&90.8$\pm$0.03\\ \hline
    \multirow{6}{*}{jaBERT-base \subword{}}&JSICK&JSICK&76.9&80.9&78.5&80.8$\pm$0.06    \\
    &JSICK (hypo-only)&JSICK&49.2&38.7&37.8&60.8$\pm$0.07\\
    &JSNLI&JSICK&57.3&63.1&48.8&48.3$\pm$0.04\\
    &&JSNLI&91.5&91.5&91.3&91.5$\pm$0.04\\
    &JSNLI+JSICK&JSICK&84.9&82.8&83.7&86.1$\pm$0.03\\ 
    &&JSNLI&91.5&91.5&91.4&91.5$\pm$0.03\\ \hline \hline
    \multicolumn{7}{c}{\textbf{Multilingual models}} \\ \hline
    \multirow{6}{*}{XLM-RoBERTa-large}
    &JSICK&JSICK&88.2&86.5&87.2&89.1$\pm$0.10\\
    &JSICK (hypo-only)&JSICK&52.0&51.2&50.2&56.1$\pm$0.09\\
    &JSNLI&JSICK&61.2&68.4&54.9&53.9$\pm$0.09\\
    &&JSNLI&94.5&94.6&94.5&94.6$\pm$0.04\\
    &JSNLI+JSICK&JSICK&89.2&89.4&89.3&90.8$\pm$0.07\\ 
    &&JSNLI&94.0&94.1&94.0&94.1$\pm$0.05\\ \hline
    \multirow{6}{*}{XLM-RoBERTa-base}
    &JSICK&JSICK&79.3&68.1&70.2&78.5$\pm$0.08\\
    &JSICK (hypo-only)&JSICK&40.3&43.7&45.5&56.8$\pm$0.06\\
    &JSNLI&JSICK&56.6&63.2&51.5&51.0$\pm$0.09\\
    &&JSNLI&92.1&92.2&92.1&92.1$\pm$0.05\\
    &JSNLI+JSICK&JSICK&85.9&86.4&86.0&88.1$\pm$0.07\\ 
    &&JSNLI&92.0&92.1&92.0&92.1$\pm$0.04\\ \hline
    \multirow{6}{*}{mBERT}
    &JSICK&JSICK&88.2&86.4&87.3&89.2$\pm$0.08\\
    &JSICK (hypo-only)&JSICK&44.7&36.2&32.9&58.8$\pm$0.09\\
    &JSNLI&JSICK&58.2&65.2&52.6&51.9$\pm$0.05\\
    &&JSNLI&91.8&92.0&91.9&92.0$\pm$0.04\\
    &JSNLI+JSICK&JSICK&87.8&87.2&87.5&89.3$\pm$0.03\\ 
    &&JSNLI&92.0&92.2&92.1&92.1$\pm$0.03\\ \hline
    \end{tabular}
    }
    \caption{Baseline results with Japanese and multilingual pre-trained language models for the NLI task with JSICK and JSNLI (\%).}
    \label{tab:rteout}
\end{table*}

\section{Baseline Experiments}
\label{sec:baseline}
\subsection{Experimental Setup}
In this study, we experimented with two Japanese pre-trained language models: Japanese BERT (\newcite{devlin-etal-2019-bert}, \bertja{}) pre-trained on Japanese Wikipedia, and Japanese RoBERTa (\newcite{RoBERTa}, jaRoBERTa) pre-trained on Japanese Wikipedia and the Japanese portion of CC-100: Monolingual Datasets from Web Crawl Data~\cite{conneau-etal-2020-unsupervised}.

For \bertja{}, 
we investigated performance differences between the BERT-base model\footnote{\url{https://huggingface.co/cl-tohoku/bert-base-japanese}} pre-trained with 17 million sentences from Wikipedia articles and the BERT-large model\footnote{\url{https://huggingface.co/cl-tohoku/bert-large-japanese}} pre-trained with 30 million sentences.
The configuration was the same as that for the original BERT model.
To check whether methods for tokenization and masked language modeling (MLM) affect model performance, we compared three settings for the BERT-base model.
In the \subword{} setting, the model processes input texts with word-level tokenization by the MeCab morphological parser with a standard Japanese dictionary IPAdic~\cite{asahara03}, followed by WordPiece subword tokenization~\cite{schuster2012japanese}.
The vocabulary size was 32,000.
In the \wholema{} setting, the subword model was trained with whole-word masking enabled for the MLM objective.
In the \charact{} setting, the model processed texts with word-level tokenization based on the IPAdic, followed by character-level tokenization.

For jaRoBERTa, we compared the performance of the base model\footnote{\url{https://huggingface.co/nlp-waseda/roberta-base-japanese}} and the large model.\footnote{\url{https://huggingface.co/nlp-waseda/roberta-large-japanese}}
The input text was segmented into words by the Japanese morphological analyzer Juman++~\cite{morita-etal-2015-morphological,tolmachev-etal-2018-juman}, and each word was tokenized using SentencePiece.\footnote{\url{https://github.com/google/sentencepiece}}

We also analyzed differences in behaviors of the Japanese and multilingual pre-trained language models.
As multilingual models, we used the multilingual BERT model (\bertm{}) trained with multilingual Wikipedia and the XLM-RoBERTa-base\footnote{\url{https://huggingface.co/xlm-roberta-base}} and XLM-RoBERTa-large\footnote{\url{https://huggingface.co/xlm-roberta-large}} models~\cite{conneau-etal-2020-unsupervised} pre-trained on CC-100 containing 100 languages. 
For \bertm{}, we used a multilingual cased model\footnote{\url{https://huggingface.co/bert-base-multilingual-cased}}, as is recommended for languages with non-Latin alphabets, like Japanese.
For each setting, we used learning rates $2e^{-5}, 3e^{-5}$, and $5e^{-5}$ and 3, 4, and 5 training epochs to tune for the best parameters.

For the NLI task, to investigate whether the size and quality of fine-tuned data affect performance, we fine-tuned pre-trained models on three types of training data: (i)~JSICK training data (5K), (ii)~JSNLI training data (533K), and (iii)~both JSICK and JSNLI training data (538K).
As mentioned in Section~\ref{sec:related}, JSNLI is a machine-translated Japanese SNLI dataset.
Since both SICK and SNLI are derived from image captions, we hypothesized that JSNLI might improve model performance on the JSICK test set.
We used four standard evaluation metrics for NLI tasks: precision (Prec), recall (Rec), macro F1-score (F1), and accuracy (Acc).
To analyze whether entailment labels are learned and predicted by referring only to hypothesis sentences, we investigated the performance of models trained on JSICK without the premise sentences. We performed five runs and present the averages below.
We also report standard deviations for the accuracy of baseline results in the NLI task.
As the baseline for the STS task, we used the Pearson correlation coefficient, Spearman correlation coefficient, and mean square error (MSE) between the prediction results for BERTScore~\cite{bert-score}, a recent BERT-based model for unsupervised STS, and the gold similarity score.

\begin{table*}[h!t]
    \centering
    \scalebox{0.73}{
    \begin{tabular}{c}
    \begin{tabular}{l|ccc|ccc|ccc|ccc|ccc|ccc}\hline
    \multirow{4}{*}{\textbf{Score}}&\multicolumn{18}{c}{\textbf{Japanese models}}\\ \cline{2-19}
    &\multicolumn{3}{c|}{\textbf{jaRoBERTa-large}}&\multicolumn{3}{c|}{\textbf{jaRoBERTa-base}}&\multicolumn{3}{c|}{\textbf{jaBERT-large}}&\multicolumn{9}{c}{\textbf{jaBERT-base}}\\
&&&&&&&&&&\multicolumn{3}{c|}{\wholema{}}&\multicolumn{3}{c|}{\charact{}}&\multicolumn{3}{c}{\subword{}}\\
    &$\gamma$&$\rho$&\textbf{MSE}&$\gamma$&$\rho$&\textbf{MSE}&$\gamma$&$\rho$&\textbf{MSE}&$\gamma$&$\rho$&\textbf{MSE}&$\gamma$&$\rho$&\textbf{MSE}&$\gamma$&$\rho$&\textbf{MSE}\\ \hline
    1-2&32.9&32.8&1.63&12.8&9.3&1.93&30.0&30.3&1.40&33.3&34.8&1.62&38.1&39.0&1.48&31.9&32.8&1.61 \\
    2-3&29.6&29.4&1.34&20.5&21.9&1.93&27.8&27.7&1.12&28.9&29.3&1.33&30.9&30.9&1.22&27.5&27.7&1.32 \\
    3-4&34.4&34.2&1.07&20.5&21.9&1.62&32.8&32.0&91.8&32.9&32.4&1.07&34.7&34.9&0.99&32.0&31.7&1.06 \\
    4-5&16.7&21.8&72.4&12.5&15.5&0.88&25.0&30.6&70.0&20.0&24.9&0.72&24.5&26.2&0.70&22.0&26.2&0.72 \\ \hline
    All&74.6&75.3&1.22&65.3&69.1&1.47&71.6&72.1&1.05&73.8&74.1&1.22&77.1&77.1&1.12&72.3&72.6&1.21\\ \hline
    \end{tabular}
    \\
    \\
    \begin{tabular}{l|ccc|ccc|ccc}\hline
    \multirow{3}{*}{\textbf{Score}}&\multicolumn{9}{c}{\textbf{Multilingual models}}\\ \cline{2-10}
    &\multicolumn{3}{c|}{\textbf{XLM-RoBERTa-large}}&\multicolumn{3}{c|}{\textbf{XLM-RoBERTa-base}}&\multicolumn{3}{c}{\textbf{\bertm{}}}\\
    &$\gamma$&$\rho$&\textbf{MSE}&$\gamma$&$\rho$&\textbf{MSE}&$\gamma$&$\rho$&\textbf{MSE}\\ \hline
    1-2&12.8&9.3&1.93&33.0&32.4&1.93&38.1&39.9&1.46\\
    2-3&20.5&21.9&1.62&28.5&28.7&1.62&30.0&29.7&1.20\\
    3-4&28.8&31.8&1.30&33.8&34.8&1.30&35.6&35.8&0.98\\
    4-5&12.5&15.5&0.88&15.6&19.2&0.88&22.2&25.3&0.69\\ \hline
    All&65.3&69.1&1.47&75.5&75.7&1.47&77.3&77.4&1.10\\ \hline
    \end{tabular}
    \end{tabular}
    }
    \caption{Baseline results from Japanese and multilingual BERTscore models} on the STS task with JSICK (\%). $\gamma$: Pearson correlation~$\times100$, $\rho$: Spearman correlation~$\times100$.
    \label{tab:stsout}
\end{table*}

\begin{table*}[h!t]
    \centering
    \scalebox{0.77}{
    \begin{tabular}{llllll}\hline
         \textbf{Model}&\textbf{Score}&\textbf{Entailment}&\textbf{Contradiction}&\textbf{Neutral}&\textbf{All}  \\ \hline
         \multirow{5}{*}{jaRoBERTa-large}&1-2&-&25.0(1/4)&100.0(647/647)&99.5(648/651)\\
         &2-3&33.3(1/3)&44.6(37/83)&97.8(1136/1162)&94.1(1174/1248)\\
         &3-4&62.4(98/157)&72.5(237/327)&89.8(991/1103)&83.6(1326/1587)\\
         &4-5&87.1(560/643)&97.9(375/383)&67.7(88/130)&88.5(1023/1156)\\ 
         &All&86.0(936/1088)&81.6(650/797)&94.1(2862/3042)&-\\        
         \hline
         \multirow{5}{*}{\bertm{}}&1-2&-&25.0 (1/4)&100.0 (647/647)&99.5 (648/651)\\
         &2-3&100.0 (3/3)&45.8 (38/83)&96.8 (1125/1162)&93.4 (1166/1248)\\
         &3-4&61.8 (97/157)&73.4 (240/327)&55.8 (615/1103)&81.9 (1299/1587)\\
         &4-5&84.6 (544/643)&96.6 (370/383)&39.2 (51/130)&86.9 (1005/1156)\\
         &All&84.7 (922/1088)&81.4 (649/797)&92.8 (2825/3042)&-\\
         \hline
    \end{tabular}
    }
    \caption{Distribution of accuracies for the JSICK NLI test set for each similarity score.}
    \label{tab:nli-rel1}
\end{table*}

\begin{table}[h!t]
    \centering
    \scalebox{0.73}{
    \begin{tabular}{llccccc}\hline
         \textbf{Model}&\textbf{Label}&\textbf{1-2}&\textbf{2-3}&\textbf{3-4}&\textbf{4-5}&\textbf{All}  \\ \hline
         &Entailment&-&-62.5&25.0&21.7&36.6\\
         jaRoBERTa&Contradiction&-83.0&13.3&15.4&13.5&39.5\\
         -large&Neutral&32.8&27.1&39.6&28.0&68.0\\
         &All&32.9&29.6&34.4&24.5&-\\ \hline
         \multirow{4}{*}{\bertm{}}&Entailment&-&-46.1&28.6&24.8&45.5\\
         &Contradiction&-82.5&18.7&20.9&13.6&45.3\\
         &Neutral&38.0&27.5&38.9&30.9&70.6\\ 
         &All&38.1&30.0&35.6&23.2&-\\ 
         \hline
    \end{tabular}
    }
    \caption{Distribution of Pearson correlations for the JSICK STS test set for each entailment label.}
    \label{tab:nli-rel2}
\end{table}

\subsection{Baseline results}
Table~\ref{tab:rteout} shows the evaluation results for NLI models.
For all models, the accuracy on JSICK is lower than that on JSNLI, indicating that JSICK poses more challenges than does JSNLI.
Since performance under the hypothesis-only setting was low, JSICK does not allow model predictions from hypotheses alone.

In the standard train/test split setting for JSICK, accuracy with
the jaRoBERTa-large model had the best performance (acc.\,$90.3\%$).
Surprisingly, multilingual models such as XLM-RoBERTa-large and \bertm{} achieved comparable accuracy
($89.1\%$ and $89.2\%$, respectively).
Among the multilingual models, the \bertm{} model had the best performance.
For the jaRoBERTa, jaBERT, and XLM-RoBERTa models, those trained on larger texts achieved higher accuracies on NLI tasks.
Among the tokenization settings for \bertja{}, whole-word masking \wholema{} provided the highest accuracy (82.4\%).
Regarding fine-tuning data, mixing the training data with the JSICK and JSNLI training sets improved model performance for the JSICK test set for all models except jaRoBERTa-large.
Since the jaRoBERTa-large model trained with a single training set (JSICK or JSNLI) already demonstrated high performance, additional training data did not improve performance.

Table~\ref{tab:stsout} shows the results from the unsupervised STS model.
Interestingly, \bertm{} achieved nearly the same high performance as did \bertja{} on the STS task.
Among different tokenization settings for \bertja{}, the character-based tokenization \charact{} produced the highest performance.
This is due to the difference between NLI and STS tasks.
Similarity scores are affected by the token overlap between two sentences, as suggested by the fact that the \lcontra{} cases tended to have higher similarity scores.
Character-based tokenization allows more precise calculations of the token overlap, and thus might be suitable for STS tasks.

\paragraph{Relevance between entailment and similarity}
We next analyze relations between entailment labels and similarity scores in cases where model predictions are difficult.
Table~\ref{tab:nli-rel1} shows a distribution of accuracies from the jaRoBERTa-large and \bertm{} NLI models fine-tuned with JSICK for each similarity score.
These results show that both Japanese and multilingual models struggled to predict entailment labels with low similarity scores, but their gold labels are \lcontra{}.
Both models also failed to predict cases where premise sentences are very similar to their hypothesis sentences but their gold labels are \lneutral{}.
Table~\ref{tab:nli-rel2} shows a distribution of Pearson correlations on the JSICK STS test set for each entailment label.
These results show that the STS models have the same trend for \lcontra{} examples as do the NLI models; the STS models failed to predict low similarity scores in cases where a premise sentence contradicts a hypothesis sentence.

\begin{table*}[h!t]
    \centering
    \scalebox{0.80}{
    \begin{tabular}{l|cccc|cccc}\hline
        \multirow{3}{*}{\textbf{Phenomena}}&\multicolumn{4}{c|}{\textbf{jaRoBERTa-large}}&\multicolumn{4}{c}{\textbf{mBERT}}\\
        &\textbf{Similarity}&\multicolumn{3}{c|}{\textbf{Entailment}}&\textbf{Similarity}&\multicolumn{3}{c}{\textbf{Entailment}}\\
        &&\textbf{I}&\textbf{N}&\textbf{N+I}&&\textbf{I}&\textbf{N}&\textbf{N+I}  \\ \hline
        Numeral (1513)&76.4&92.7&\lowacc{50.2}&89.4&80.0&91.9&\lowacc{45.8}&91.2\\
        Negation (1140)&80.4&93.9&\lowacc{49.5}&93.5&82.8&94.6&\lowacc{45.6}&93.4\\
        Quantification (744)&\lowacc{73.9}&91.7&\lowacc{51.3}&89.6&78.1&90.9&\lowacc{49.6}&90.7\\
        Passive (695)&\lowacc{73.1}&91.1&60.0&90.3&\lowacc{77.1}&90.4&57.7&89.4\\ 
        Anaphora (700)&\lowacc{73.7}&\lowacc{89.7}&\lowacc{53.1}&\lowacc{86.4}&\lowacc{77.0}&89.3&53.1&89.4\\ 
        Conjunction (640)&78.8&91.6&56.2&\lowacc{84.1}&79.5&90.8&55.2&90.9\\
        Disjunction (428)&\lowacc{73.5}&\lowacc{90.0}&\lowacc{49.3}&\lowacc{84.5}&79.2&\lowacc{89.0}&\lowacc{46.7}&\lowacc{89.0}\\
        Modal (69)&90.4&91.3&60.9&88.3&92.7&94.2&53.6&92.8\\
        Additive particle (13)&77.1&\lowacc{84.6}&53.8&\lowacc{79.2}&\lowacc{64.9}&\lowacc{76.9}&\lowacc{30.8}&\lowacc{76.9}\\
        \hline
        All&74.6&90.3&53.8&87.9&77.3&89.2&51.9&89.3\\
         \hline
    \end{tabular}
    }
    \caption{Results on JSICK for each linguistic tag (\%). We evaluated NLI models for accuracy and STS models with the Pearson correlation$\times100$. N: JSNLI-train, I: JSICK-train, N+I: JSNLI+JSICK-train. Accuracies lower than the overall model accuracy are indicated in red.}
    \label{tab:semrteout}
\end{table*}

\begin{table}[h!t]
    \centering
    \scalebox{0.7}{
    \begin{tabular}{l|cccc|ccc}\hline
    \multirow{2}{*}{\textbf{Language}}&\multicolumn{4}{c|}{\textbf{NLI task}}&\multicolumn{3}{c}{\textbf{STS task}}\\
    &\textbf{Prec}&\textbf{Rec}&\textbf{macro-F1}&\textbf{Acc}&$\gamma$&$\rho$&\textbf{MSE}\\ \hline
    En&87.2&84.9&86.0&86.6&59.9&56.1&0.98\\
    Nl&86.5&83.4&85.3&86.2&57.8&54.6&0.95\\
    Br&85.4&83.1&84.2&85.0&61.3&57.2&0.97\\
    Ja (L-En)&84.1&82.9&83.0&85.2&62.3&60.7&1.08 \\ 
    Ja (L-Ja)&88.2&86.4&87.3&89.2&77.3&77.4&1.11 \\ \hline
    \end{tabular}
    }
    \caption{Baseline results from \bertm{} on different languages in the SICK test set (\%). $\gamma$: Pearson correlation~$\times100$, $\rho$: Spearman correlation~$\times100$. Ja (L-En) indicates evaluation results with gold labels from the original SICK test set.}
    \label{tab:multibert1}
\end{table}

\begin{table*}[h!t]
\centering
    \scalebox{0.85}{
    \begin{tabular}{cl|cccc}
    && \multicolumn{3}{c}{\textbf{Prediction En}} \\
    & & \multicolumn{1}{c}{\textbf{E}} & \multicolumn{1}{c}{\textbf{C}} & \multicolumn{1}{c}{\textbf{N}} & \textbf{Rec} \\ \hline
    \multirow{3}{*}{\rotatebox{90}{\textbf{Gold}}}
    & \textbf{E} & 1151 & 2 & 261 & 81.4\% \\
    & \textbf{C} & 16 & 601 & 103 & 83.5\% \\
    & \textbf{N} & 229 & 51 & 2513 & 90.0\% \\ 
    & \textbf{Prec} & \multicolumn{1}{c}{82.4\%} & \multicolumn{1}{c}{91.9\%} & \multicolumn{1}{c}{87.3\%} & \\
    \end{tabular}
    \hspace{1ex}
    \begin{tabular}{cccc}
    \multicolumn{3}{c}{\textbf{Prediction Nl}} \\
    \multicolumn{1}{c}{\textbf{E}} & \multicolumn{1}{c}{\textbf{C}} & \multicolumn{1}{c}{\textbf{N}} & \textbf{Rec} \\ \hline
    1083 & 5 & 316 & 77.1\% \\
    17 & 592 & 103 & 83.1\% \\
    217 & 67 & 2506 & 89.8\% \\ 
    \multicolumn{1}{c}{82.2\%} & \multicolumn{1}{c}{89.2\%} & \multicolumn{1}{c}{85.7\%} & \\
    \end{tabular}
    \hspace{1ex}
    \begin{tabular}{cccc}
    \multicolumn{3}{c}{\textbf{Prediction Br}} \\
    \multicolumn{1}{c}{\textbf{E}} & \multicolumn{1}{c}{\textbf{C}} & \multicolumn{1}{c}{\textbf{N}} & \multicolumn{1}{c}{\textbf{Rec}} \\ \hline
    1094 & 5 & 305 & 77.9\% \\
    23 & 586 & 103 & 82.3\% \\
    238 & 63 & 2489 & 89.2\% \\ 
    \multicolumn{1}{c}{80.7\%} & \multicolumn{1}{c}{89.6\%} & \multicolumn{1}{c}{85.9\%} & \\
    \end{tabular}
}
    \scalebox{0.85}{
    \begin{tabular}{cl|cccc}
    &&\multicolumn{3}{c}{\textbf{Prediction Ja (L-En)}} \\
    && \multicolumn{1}{c}{\textbf{E}} & \multicolumn{1}{c}{\textbf{C}} & \multicolumn{1}{c}{\textbf{N}} & \multicolumn{1}{c}{\textbf{Rec}} \\ \hline    
    \multirow{3}{*}{\rotatebox{90}{\textbf{Gold}}}
    & \textbf{E} & 972 & 1 & 115 & 89.3\% \\
    & \textbf{C} & 16 & 574 & 207 & 72.0\% \\
    & \textbf{N} & 315 & 73 & 2654 & 87.2\% \\
    & \textbf{Prec} &\multicolumn{1}{c}{74.6\%} & \multicolumn{1}{c}{88.6\%} & \multicolumn{1}{c}{89.2\%} & \\
    \end{tabular}
    \hspace{1em}
    \begin{tabular}{cccc}
    \multicolumn{3}{c}{\textbf{Prediction Ja (L-Ja)}} \\
    \multicolumn{1}{c}{\textbf{E}} & \multicolumn{1}{c}{\textbf{C}} & \multicolumn{1}{c}{\textbf{N}} & \multicolumn{1}{c}{\textbf{Rec}} \\ \hline    
    922 & 12 & 154 & 84.7\% \\
    11 & 649 & 137 & 81.4\% \\
    144 & 73 & 2825 & 92.9\% \\
    \multicolumn{1}{c}{85.6\%} & \multicolumn{1}{c}{88.4\%} & \multicolumn{1}{c}{90.7\%} & \\
    \end{tabular}
    }
    \caption{Confusion matrices for mBERT on different languages in the SICK NLI test set. \textbf{Rec} and \textbf{Prec} indicate ``Recall'' and ``Precision'', respectively.}
    \label{table:confusion}
\end{table*}

\begin{table*}[h!t]
    \centering
    \scalebox{0.80}{
    \begin{tabular}{l|ccccc|ccccc}\hline
    \multirow{2}{*}{\textbf{Phenomenon}}&\multicolumn{5}{c|}{\textbf{NLI task}}&\multicolumn{5}{c}{\textbf{STS task}}\\
    &\textbf{En}&\textbf{Nl}&\textbf{Br}&\textbf{Ja (L-En)}&\textbf{Ja (L-Ja)} &\textbf{En}&\textbf{Nl}&\textbf{Br}&\textbf{Ja (L-En)}&\textbf{Ja (L-Ja)}\\ \hline
     Numeral (1513)&87.9&\lowacc{85.8}&\lowacc{85.1}&88.2&91.9&66.2&63.3&65.9&65.3&80.0\\
     Negation (1140)&90.3&89.3&88.8&90.7&94.6&61.3&59.0&60.9&\lowacc{58.0}&82.8\\
     Quantification (744)&\lowacc{86.0}&\lowacc{85.4}&\lowacc{83.8}&87.9&90.9&62.3&58.7&\lowacc{60.3}&\lowacc{58.7}&78.1\\
     Passive (695)&87.6&\lowacc{84.4}&86.2&87.3&90.4&62.7&58.7&64.2&68.6&\lowacc{77.1}\\ 
     Anaphora (700)&\lowacc{85.3}&87.4&85.4&87.0&89.3&\lowacc{59.8}&\lowacc{55.7}&\lowacc{60.8}&\lowacc{60.4}&\lowacc{77.0}\\ 
     Conjunction (640)&\lowacc{84.7}&\lowacc{85.9}&85.8&86.2&90.8&64.7&63.3&65.4&64.9&79.5\\
     Disjunction (428)&\lowacc{85.5}&\lowacc{83.9}&\lowacc{83.8}&87.6&\lowacc{89.0}&65.3&59.9&62.9&65.1&79.2\\
     Modal (69)&92.8&95.7&88.4&89.9&94.2&85.5&81.3&85.7&84.3&92.7\\
     Additive particle (13)&\lowacc{61.5}&\lowacc{69.2}&\lowacc{53.8}&\lowacc{53.8}&\lowacc{76.9}&\lowacc{44.2}&\lowacc{24.5}&\lowacc{58.7}&\lowacc{39.9}&\lowacc{64.9}\\ \hline
     All&86.6&86.2&85.0&85.2&89.2&55.9&57.8&61.3&62.3&77.3 \\
     \hline
    \end{tabular}
    }
    \caption{Comparison of \bertm{} performance for different languages in the SICK test set for each linguistic tag (\%). Accuracies lower than the overall model accuracy are indicated in red. Ja (L-En) indicates evaluation results with gold labels from the original SICK test set.}
    \label{tab:multibert2}
\end{table*}

\paragraph{Linguistic phenomena}
Table~\ref{tab:semrteout} shows evaluation results for the jaRoBERTa-large model and the \bertm{} model for each linguistic tag. 
Regarding the jaRoBERTa-large model performance, there is little difference for each linguistic phenomenon, and accuracy for examples involving anaphora, disjunction, and additive particle was comparatively low in the NLI task.
For the STS task, Pearson correlations for examples involving quantification and passive were slightly low.

Regarding differences between training data, models fine-tuned with JSICK performed better for almost all linguistic phenomena than did those fine-tuned with JSNLI.
Furthermore, adding the JSNLI training set to the JSICK training set did not improve the model performance on most linguistic phenomena.
This suggests that training data quality is more critical for learning linguistic phenomena than is quantity.

\subsection{Comparison with other languages}
\label{subsec:multi}
We next compare the performance of \bertm{} on the JSICK dataset with that on the original English SICK dataset (SICK-EN), the Portguese SICK-BR dataset~\cite{real2018sickBR}, and the Dutch SICK-NL dataset~\cite{wijnholds2021sicknl}.
Since gold labels for the JSICK datasets differ from those for the SICK-EN and SICK-NL datasets (SICK-NL uses the same labels as SICK-EN), we also evaluated \bertm{} while assuming JSICK gold labels to be the same as those for SICK-EN.
Table~\ref{tab:multibert1} shows the baseline results for \bertm{} with different languages in the SICK test set.
For both STS and NLI tasks, \bertm{} performance was relatively higher for Japanese SICK than that for the other datasets.

Table~\ref{table:confusion} shows confusion matrices for multilingual BERT models on different languages of the SICK NLI test set.
Comparing across languages, \bertm{} performance for contradiction cases was lower in Japanese.
Table~\ref{tab:multibert2} compares
\bertm{} performance on different languages of the SICK test set for each Japanese linguistic tag.
Note that since linguistic phenomena manifest differently by language, Table~\ref{tab:multibert2} shows only an approximated comparison of linguistic phenomena.
For the STS task, there was little difference among languages, but performance tended to be lower for problems involving additive particles and anaphora.
Performance for problems involving additive particles was also low in the NLI task.
Performance for problems involving disjunction was low for all languages.

The results of our experiments suggest that multilingual BERT models achieved high performance on SICK across languages.
However, the results related to multilingual SICK for each linguistic tag indicate room for improvement regarding the use of multilingual models to capture anaphora, disjunction, and additive particles.
Moreover, it remains to be investigated
whether pre-trained language models are sensitive to
compositional aspects of inference,
such as word order and case marking in Japanese.
In the next section, we describe the extent to which language models capture word order and case particles, phenomena that are characteristic of Japanese.

\begin{table*}[h!t]
    \centering

\scalebox{0.80}{
\begin{tabular}{llllllll} \hline
\premise{}:
&\NPone{小さな}&\NPone{女の子}&\Nom{が}&\NPtwo{女性}&\Acc{を}&\TVo{見て}&\TVo{いる}\\
&\NPone{little}&\NPone{girl}&\Nom{Nom}&\NPtwo{woman}&\Acc{Acc}&\TVo{looking}&\TVo{is}\\
&\multicolumn{6}{l}{`A little girl is looking at a woman'}&\spair{}: \lentail{}\\ \hline
\premord{}:
&\NPtwo{女性}&\Acc{を}&\NPone{小さな}&\NPone{女の子}&\Nom{が}&\TVo{見て}&\TVo{いる}\\
&\NPtwo{woman}&\Acc{Acc}&\NPone{little}&\NPone{girl}&\Nom{Nom}&\TVo{looking}&\TVo{is}\\
&\multicolumn{6}{l}{`A little girl is looking at a woman'}&\scrampair{}: \lentail{}\\ \hline
\premcase{}:
&\NPone{小さな}&\NPone{女の子}&\Acc{を}&\NPtwo{女性}&\Nom{が}&\TVo{見て}&\TVo{いる}\\
&\NPone{little}&\NPone{girl}&\Acc{Acc}&\NPtwo{woman}&\Nom{Nom}&\TVo{looking}&\TVo{is}\\
&\multicolumn{6}{l}{`A woman is looking at a little girl'}&\swappair{}: \lneutral{}\\ \hline
\premdel{}:
&\NPone{小さな}&\NPone{女の子}&\NPtwo{女性}&\TVo{見て}&\TVo{いる}\\
&\NPone{little}&\NPone{girl}&\NPtwo{woman}&\TVo{looking}&\TVo{is}\\
&\multicolumn{6}{l}{`A woman is looking at a little girl'}&\delpair{}: \lneutral{}\\ \hline
\hypothesis{}:
&\NPone{女の子}&\Nom{が}&\NPtwo{女性}&\Acc{を}&\TVo{見て}&\TVo{いる}\\
&\NPone{girl}&\Nom{Nom}&\NPtwo{woman}&\Acc{Acc}&\TVo{looking}&\TVo{is}\\
&\multicolumn{6}{l}{`A girl is looking at a woman'}\\ \hline
\end{tabular}
}
\caption{Evaluation settings for the JSICK stress-test dataset.}
\label{tab:setstress}
\end{table*}

\begin{table*}[h!t]
\centering
\scalebox{0.77}{
\begin{tabular}{c}
\begin{tabular}{l|cccc|cc}\hline
\multirow{2}{*}{\textbf{Type}}
&\multicolumn{4}{c|}{\textbf{Japanese models}}&
\multirow{2}{*}{\textbf{Human}}&\multirow{2}{*}{\textbf{Natural}}\\ 
& \textbf{jaRoBERTa-large} & \textbf{jaRoBERTa-base} &
\textbf{\bertja{}-large} &\textbf{\bertja{}-base}~\wholema{}  \\ \hline
Case-scrambling & 98.9 & 97.5 & 92.4 & 97.3 & 93.3 & 97.3 \\
Part-swapping & 99.0 & 97.0 & 92.5 & 98.3 & 66.7 & 23.0                     \\
Part-deleting & 98.3 & 95.4 & 92.3 & 91.7 & 85.3 & 6.0 \\ \hline                     
\end{tabular}
\\ \\
\begin{tabular}{l|ccc}\hline
\multirow{2}{*}{\textbf{Type}}
& \multicolumn{3}{c}{\textbf{Multilingual models}} \\
&\textbf{XLM-RoBERTa-large}&\textbf{XLM-RoBERTa-base}&\textbf{\bertm{}}\\ \hline
Case-scrambling & 98.4 & 94.5 & 98.4  \\ 
Part-swapping & 98.7 & 94.7 & 99.4 \\ 
Part-deleting & 98.0 & 92.0 & 97.2 \\ \hline 
\end{tabular}
\end{tabular}
}
\caption{Comparison between model and human predictions for entailment labels in the JSICK stress set that are the same as those for the original test set (\%). \textbf{Natural} indicates human-rated results for naturalness (acceptability).}
\label{tab:stresshuman}
\end{table*}

\begin{table*}[h!t]
\centering
\scalebox{0.72}{
\begin{tabular}{lll|cccc|c}\hline
\multirow{2}{*}{\textbf{Order}}                  &\multirow{2}{*}{\textbf{Model}}&\multirow{2}{*}{\textbf{Type}}&\multicolumn{4}{c|}{\textbf{NLI task}}&\multirow{2}{*}{\textbf{STS task}} \\ 
&&& \textbf{Yes}  & \textbf{No}   & \textbf{Unk}  &\textbf{All}&\\ \hline
\multirow{21}{*}{ga-de} 
                       &\multirow{3}{*}{jaRoBERTa-large}& Case-scrambling &98.8&100.0&99.5&99.4&98.8  \\
                       &                        &
                       Part-swapping &97.2&99.1&99.8&99.1&97.0   \\
                       &                        &
                       Part-deleting &95.2&99.1&99.1&98.1&92.6   \\ 
                       \cline{2-8}
                       &\multirow{3}{*}{jaRoBERTa-base}& Case-scrambling &98.4&96.6&98.0&97.9&95.0  \\
                       &                        &
                       Part-swapping &97.6&97.4&96.9&97.1&99.1   \\
                       &                        &
                       Part-deleting &95.6&95.7&95.0&95.2&91.3   \\ 
                       \cline{2-8}
                       &\multirow{3}{*}{\bertja{}-large}& Case-scrambling &84.3&87.2&94.2&90.9&98.6  \\
                       &                        &
                       Part-swapping &85.1&88.0&94.1&91.1&95.8   \\
                       &                        &
                       Part-deleting &83.5&86.3&94.7&90.9&94.1   \\ 
                       \cline{2-8}
                       &\multirow{3}{*}{\bertja{}-base \wholema{}}& Case-scrambling &95.5&100.0&97.2&97.1&99.2  \\
                       &                        &
                       Part-swapping &98.8&100.0&99.4&99.3&96.8   \\
                       &                        &
                       Part-deleting &91.5&96.6&95.2&94.4&92.4 \\ 
                       \cline{2-8}
                       & \multirow{3}{*}{XLM-RoBERTa-large} & Case-scrambling &  98.8&98.3&99.2&99.0&89.4\\
                       &                        & 
                       Part-swapping &100.0&98.3&99.2&99.3&94.2  \\ 
                       &                        &
                       Part-deleting &97.2&97.4&98.4&98.0&86.8 \\
                       \cline{2-8}
                       & \multirow{3}{*}{XLM-RoBERTa-base} & Case-scrambling &89.1&93.2&96.9&94.5&97.7\\
                       &                        & 
                       Part-swapping &85.5&90.6&97.3&93.6&96.9  \\ 
                       &                        &
                       Part-deleting &79.4&88.9&95.5&90.7&88.1 \\
                       \cline{2-8}
                       & \multirow{3}{*}{\bertm{}} & Case-scrambling &97.2&97.4&98.6&98.1&98.0  \\
                       &                        & Part-swapping &100.0&99.1&100.0&99.9&98.3  \\ 
                       &                        &
                       Part-deleting &94.8&99.1&98.4&97.6&96.4 \\
                       \hline
\multirow{21}{*}{ga-ni} 
                       &\multirow{3}{*}{jaRoBERTa-large}& Case-scrambling &97.2&96.0&99.2&98.4&98.6  \\
                       &                        &
                       Part-swapping &97.7&96.0&99.2&98.5&97.0   \\
                       &                        &
                       Part-deleting &96.0&96.0&98.7&97.7&93.6   \\ 
                       \cline{2-8}
                       &\multirow{3}{*}{jaRoBERTa-base}& Case-scrambling &98.3&97.0&97.5&97.6&96.7  \\
                       &                        &
                       Part-swapping &98.9&97.0&97.1&97.5&94.6   \\
                       &                        &
                       Part-deleting &97.2&95.0&94.9&95.4&91.3   \\ 
                       \cline{2-8}
                       &\multirow{3}{*}{\bertja{}-large}& Case-scrambling &90.3&87.0&93.7&92.1&97.7  \\
                       &                        &
                       Part-swapping &91.5&89.0&93.8&92.7&95.2   \\
                       &                        &
                       Part-deleting &91.5&86.0&93.5&92.1&94.1   \\ 
                       \cline{2-8}
                       & \multirow{3}{*}{\bertja{}-base \wholema{}}    & Case-scrambling &97.2&97.0&96.4&96.6&98.2  \\
                       &                        &
                       Part-swapping &99.4&99.0&98.5&98.7&96.0  \\
                       &                        &
                       Part-deleting &93.2&94.0&92.3&92.7&92.0 \\
                       \cline{2-8}
                       & \multirow{3}{*}{XLM-RoBERTa-large} & Case-scrambling &99.4&96.0&99.2&98.9&84.8  \\
                       &                        & 
                       Part-swapping & 98.3&95.0&99.0&98.4&93.5 \\ 
                       &                        &
                       Part-deleting &98.3&95.0&99.0&98.4&86.9 \\
                       \cline{2-8}
                       & \multirow{3}{*}{XLM-RoBERTa-base} & Case-scrambling &94.9&82.0&96.0&94.0&93.0\\
                       &                        & 
                       Part-swapping &92.0&84.0&97.9&94.8&91.9  \\ 
                       &                        &
                       Part-deleting &91.5&81.0&94.6&92.2&88.3 \\
                       \cline{2-8}
                       & \multirow{3}{*}{\bertm{}} & Case-scrambling &98.9&94.0&99.6&98.7&95.7  \\
                       &                        & Part-swapping &98.9&97.0&97.0&99.6&98.4 \\ 
                       &                        &
                       Part-deleting &97.7&94.0&98.8&98.0&96.0 \\ 
                       \hline
\multirow{21}{*}{ga-o} 
                       &\multirow{3}{*}{jaRoBERTa-large}& Case-scrambling &98.3&97.7&99.5&99.0&98.5  \\
                       &                        &
                       Part-swapping &98.6&98.9&99.6&99.3&95.6   \\
                       &                        &
                       Part-deleting &98.3&98.1&99.1&98.8&92.6   \\ 
                       \cline{2-8}
                       &\multirow{3}{*}{jaRoBERTa-base}& Case-scrambling &96.3&96.6&97.8&97.3&91.4  \\
                       &                        &
                       Part-swapping &95.8&95.5&97.3&96.7&97.2   \\
                       &                        &
                       Part-deleting &95.5&95.1&95.5&95.4&95.9   \\ 
                       \cline{2-8}
                       &\multirow{3}{*}{\bertja{}-large}& Case-scrambling &89.3&92.4&95.2&93.5&96.9  \\
                       &                        &
                       Part-swapping &88.4&92.8&94.9&93.2&93.5   \\
                       &                        &
                       Part-deleting &89.8&92.0&94.8&93.3&90.6   \\ 
                       \cline{2-8}
                       &\multirow{3}{*}{\bertja{}-base \wholema{}}    & Case-scrambling &96.1 & 98.1 & 96.1 & 96.4&97.8  \\
                       &                        &
                       Part-swapping & 98.3 & 98.9 & 99.1 & 98.9 &94.8  \\
                       &                        &
                       Part-deleting & 92.4 & 93.6 & 92.5 & 92.6 & 90.0\\ 
                       \cline{2-8}
                       & \multirow{3}{*}{XLM-RoBERTa-large} & Case-scrambling & 97.2&96.6&98.2&97.7&87.5 \\
                       &                        & 
                       Part-swapping &98.3&97.7&98.6&98.4&94.5  \\ 
                       &                        &
                       Part-deleting &96.6&96.2&98.7&97.8&86.8 \\
                       \cline{2-8}
                       & \multirow{3}{*}{XLM-RoBERTa-base} & Case-scrambling &90.4&90.5&97.3&94.8&96.5\\
                       &                        & 
                       Part-swapping &92.4&92.8&96.8&95.3&94.9 \\ 
                       &                        &
                       Part-deleting &86.4&89.0&95.5&92.6&85.1 \\
                       \cline{2-8}
                       & \multirow{3}{*}{\bertm{}} & Case-scrambling &99.2&97.7&98.3&98.4&95.5  \\
                       &                        & Part-swapping &99.2&98.5&99.4&99.2&97.9  \\ 
                       &                        &
                       Part-deleting &95.5&94.7&97.5&96.6&94.1  \\
                       \hline
\end{tabular}
}
\caption{Percentage of predictions for the JSICK stress-test dataset that are the same as those for the original test set for each case particle (\%). ``Yes'', ``No'', and ``Unk'' indicate accuracies on \lentail{}, \lcontra{}, and \lneutral{} examples, respectively. For the STS task, we calculated the Pearson correlation between predictions for the original pairs and those for the rephrased pairs.}
\label{tab:stress2}
\end{table*}

\section{JSICK Stress Test}
\label{sec:stress}

\subsection{Evaluation setting}
Japanese grammar allows both subject--object--verb and object--subject--verb orders, with the former usually taken as the basic word order
and the latter derived by a \textit{scrambling} operation~\cite{hoji1985,saito1985some}.
Instead of word order, postpositional case particles function as case markers.
For example, the case particles \textit{ga}, \textit{ni}, \textit{o} represent the nominative, dative, and accusative cases, respectively.
The JSICK test set contains 1666, 797, and 1006 premise--hypothesis sentence pairs \spair{} whose premise sentences \premise{} include basic word orders involving \textit{ga-o} (nominative--accusative), \textit{ga-ni} (nominative--dative), and \textit{ga-de} (nominative--instrumental/locative) relations, respectively.
By transforming the syntactic structures of these pairs, we created a JSICK stress-test dataset involving word scrambling and particle-swapping to analyze whether models correctly capture the free-order property of Japanese.

Consider the examples of \spair{} pairs in Table~\ref{tab:setstress}, whose gold labels are
\lentail{}.
We first provide a scrambled pair \scrampair{}, where the word order of the premise sentence \premise{} is scrambled into \textit{o-ga}, \textit{ni-ga}, or \textit{de-ga} order.
Since the meaning of the sentence \premord{} is the same as that of the original sentence \premise{}, the semantic relatedness of the scrambled pair \scrampair{} should be the same as those of \spair{}.
To analyze whether models consider Japanese case particles when predicting entailment labels and similarity scores, we use a rephrased pair \swappair{}, where the only case particles in the premise \premise{} are swapped, and a rephrased pair \delpair{}, where the case particles in \premise{} are deleted.
Since these transformations affect case relations in premise sentences, the meanings of sentences \premcase{} and \premdel{} should differ from those of the original sentence \premise{}.
The semantic relatedness of the rephrased pairs \swappair{} and \delpair{} should thus also be changed.
If a model has generalized word order and case particles, it should consistently predict the same labels for both \spair{} and \scrampair{}.
Moreover, the model should change the labels for \swappair{} and \delpair{} to \lneutral{}.
We therefore checked the extent to which models changed predictions for \scrampair{}, \swappair{}, and \delpair{} pairs as compared with those for \spair{}.

To rephrase premise sentences, we first parse the sentences using
the Japanese constituency parser depccg~\cite{yoshikawa:2017acl},
transform the parse trees using Tsurgeon~\cite{levy-andrew-2006-tregex}, and produce their surface strings as rephrased premise sentences.
For the JSICK stress set, we evaluated four Japanese pre-trained language models
(jaRoBERTa-large, jaRoBERTa-base, jaBERT-large, and jaBERT-base (\textsc{whole}))
and three multilingual models (XLM-RoBERTa-large, XLM-RoBERTa-base, and mBERT) fine-tuned with the JSICK training set.
For the NLI task, we also used crowdsourcing to collect human judgments on a subset of the JSICK stress set.
We asked the same annotators as those assigning entailment labels for the JSICK dataset to also annotate entailment labels for the JSICK stress set.
We also asked the annotators whether each sentence pair was natural, meaning both premise and hypothesis sentences were grammatically correct.
Note that we asked them to judge entailment labels even for unnatural sentences.
We selected 100 examples for each of three rephrase types and three case particle types, for 900 inference problems in total.

\subsection{Results}
Table~\ref{tab:stresshuman} compares percentages of model and human predictions for entailment labels for the JSICK stress-test dataset that are the same as those for the original JSICK test set.
We also show the results for the human naturalness (acceptability) rating task.
While humans predicted the same labels for the scrambled examples, they changed their labels for examples where only the case particles were swapped or deleted.
Interestingly, humans tend to change more predictions for those examples where only the case particles are swapped than for those where the case particles are deleted, but the acceptability rate for the former was much higher than that for the latter.
One reason for this is that Japanese case particles can be dropped under adjacency to become a verb~\cite{saito1985some}, in which case humans can complement the dropped case particles.
On the other hand, the models predicted nearly the same labels for \scrampair{} pairs as for \spair{} pairs.
In addition, the predicted labels for \swappair{} pairs and \delpair{} remained almost the same as those for the original \spair{} pairs.

To investigate details of the model performance, we confirmed the percentage of predictions for each case particle in the JSICK stress-test dataset that are the same as those predicted for
the original test set, as shown in Table~\ref{tab:stress2}.
In the NLI task, all pre-trained language models predicted nearly the same labels even when the case particle is swapped or deleted, regardless of model type and the kind of case particle.
These results indicate that the models predict entailment labels without considering word order and case particles.
Similarly, for the STS task, Pearson correlations for \scrampair{} and \swappair{} pairs were nearly the same as those for the original \spair{} pairs.
Pearson correlations for \delpair{} pairs were a little lower than those for the original \spair{} pairs, because the deletion of case particles in \premdel{} decreases word overlap between \premdel{} and \hypothesis{}.

\paragraph{Augmenting training data for sensitivity to case particles}
To analyze whether data augmentation improves the model behavior for word order and case particles, we rephrased a small subset of the training set in three ways, creating (i)~data where the NP argument is scrambled but its gold label is the same as the original, (ii)~data where the case particle is deleted, and its gold label is set randomly, and (iii)~data where only the case particle is scrambled and its gold label is set randomly.

We added 300 examples of each data type.
These additional training data play a role in exposing models to three cues: (i)~the order of NP arguments does not change entailment labels for sentence pairs, (ii)~the existence of a case particle, and (iii)~that its position can change entailment labels.
Table~\ref{tab:dataaug} shows the percentage of predictions by the jaRoBERTa-large NLI model that are the same as those for the original JSICK when a subset of rephrased examples is added to the training set.
As that table shows, data augmentation changed model predictions on examples where the case particle is swapped or deleted.
This indicates that although the NLI model does not implicitly learn case particles during pre-training and fine-tuning, a small amount of data augmentation to learn word order and case particles can improve the model sensitivity to case particles.

\begin{table}[h!t]
\centering
\scalebox{0.77}{
\begin{tabular}{l|rr}\hline
\textbf{Type}& \multicolumn{1}{l}{\textbf{jaRoBERTa-l}} & \multicolumn{1}{l}{\textbf{jaRoBERTa-l+aug}} \\ \hline
Case-scrambling & 98.9                        &97.7                        \\
Part-swapping & 99.0                        &69.2                       \\
Part-deleting   & 98.3                        &69.1  \\ 
\hline
\end{tabular}
}
\caption{Percentages of model predictions that are the same as those for the original JSICK when a subset of rephrased examples is added to the training set to learn word order and case particles. \textbf{jaRoBERTa-l+aug} shows the result with data augmentation.}
\label{tab:dataaug}
\end{table}

\section{Conclusion}
We introduced JSICK, a Japanese standard NLI/STS dataset, by manually translating English SICK into Japanese and re-annotating its gold labels.
In baseline experiments,
we compared the performance of various pre-trained Japanese language models on JSICK.
While the Japanese RoBERTa-large model achieved state-of-the-art performance, the performance of multilingual pre-trained language models achieved comparable results.
Experiments with multilingual models on SICK datasets
in different languages, including JSICK, showed that the performance of multilingual models was relatively low on inference problems involving anaphora, disjunction, and additive particles.

Furthermore, to investigate the extent to which Japanese and multilingual pre-trained language models are sensitive to word order and case particles,
we provided a JSICK stress-test dataset involving word scrambling and particle-swapping from JSICK.
The results from that dataset suggest that both Japanese and multilingual models do not consider word order and case particles when making predictions for Japanese NLI/STS tasks.
These are novel findings that are not obtainable from other datasets, including SICK in English and other languages.
Overall, the results suggest large room for improvement of both Japanese and multilingual pre-trained language models regarding their sensitivity to flexible word order and the representations of case particles.
Further improvements might be obtained
by more adequate representations of Japanese vocabularies in multilingual pre-trained language models~\cite{chung-etal-2020-improving,rust-etal-2021-good}.
We believe our dataset will be useful in future research for realizing more advanced models capable of appropriately performing multilingual compositional inference.

\section*{Acknowledgements}
We thank the anonymous reviewers and the Action Editor for
helpful comments and suggestions that improved this paper.
We also thank Daisuke Kawahara and Tomohide Shibata for helpful advice on the experimental settings of Japanese RoBERTa and BERT models.
This work was supported by JSPS KAKENHI Grant Number JP20K19868, JST, PRESTO Grant Number JPMJPR21C8, and JST, CREST Grant Number JPMJCR2114, Japan.

\bibliography{tacl2018}

\begin{thebibliography}{67}
\expandafter\ifx\csname natexlab\endcsname\relax\def\natexlab#1{#1}\fi

\bibitem[{Agirre et~al.(2016)Agirre, Banea, Cer, Diab, Gonzalez-Agirre,
  Mihalcea, Rigau, and Wiebe}]{agirre-etal-2016-semeval}
Eneko Agirre, Carmen Banea, Daniel Cer, Mona Diab, Aitor Gonzalez-Agirre, Rada
  Mihalcea, German Rigau, and Janyce Wiebe. 2016.
\newblock {S}em{E}val-2016 task 1: Semantic textual similarity, monolingual and
  cross-lingual evaluation.
\newblock In \emph{Proceedings of the 10th International Workshop on Semantic
  Evaluation ({S}em{E}val-2016)}, pages 497--511.

\bibitem[{Amirkhani et~al.(2020)Amirkhani, Jafari, Amirak, Pourjafari, Jahromi,
  and Kouhkan}]{amirkhani2020farstail}
Hossein Amirkhani, Mohammad~Azari Jafari, Azadeh Amirak, Zohreh Pourjafari,
  Soroush~Faridan Jahromi, and Zeinab Kouhkan. 2020.
\newblock Fars{T}ail: A {P}ersian natural language inference dataset.
\newblock \emph{CoRR}, cs.CL/2009.08820.
\newblock Version 1.

\bibitem[{Asahara and Matsumoto(2003)}]{asahara03}
Masayuki Asahara and Yuji Matsumoto. 2003.
\newblock ipadic version 2.7.0 {U}ser's {M}anual.
\newblock Nara Institute of Science and Technology.

\bibitem[{Bowman et~al.(2015)Bowman, Angeli, Potts, and
  Manning}]{bowman-etal-2015-large}
Samuel~R. Bowman, Gabor Angeli, Christopher Potts, and Christopher~D. Manning.
  2015.
\newblock A large annotated corpus for learning natural language inference.
\newblock In \emph{Proceedings of the 2015 Conference on Empirical Methods in
  Natural Language Processing}, pages 632--642.

\bibitem[{Cer et~al.(2017)Cer, Diab, Agirre, Lopez-Gazpio, and
  Specia}]{cer-etal-2017-semeval}
Daniel Cer, Mona Diab, Eneko Agirre, I{\~n}igo Lopez-Gazpio, and Lucia Specia.
  2017.
\newblock {S}em{E}val-2017 task 1: Semantic textual similarity multilingual and
  crosslingual focused evaluation.
\newblock In \emph{Proceedings of the 11th International Workshop on Semantic
  Evaluation ({S}em{E}val-2017)}, pages 1--14.

\bibitem[{Chung et~al.(2020)Chung, Garrette, Tan, and
  Riesa}]{chung-etal-2020-improving}
Hyung~Won Chung, Dan Garrette, Kiat~Chuan Tan, and Jason Riesa. 2020.
\newblock Improving multilingual models with language-clustered vocabularies.
\newblock In \emph{Proceedings of the 2020 Conference on Empirical Methods in
  Natural Language Processing (EMNLP)}, pages 4536--4546.

\bibitem[{Conneau et~al.(2020)Conneau, Khandelwal, Goyal, Chaudhary, Wenzek,
  Guzm{\'a}n, Grave, Ott, Zettlemoyer, and
  Stoyanov}]{conneau-etal-2020-unsupervised}
Alexis Conneau, Kartikay Khandelwal, Naman Goyal, Vishrav Chaudhary, Guillaume
  Wenzek, Francisco Guzm{\'a}n, Edouard Grave, Myle Ott, Luke Zettlemoyer, and
  Veselin Stoyanov. 2020.
\newblock Unsupervised cross-lingual representation learning at scale.
\newblock In \emph{Proceedings of the 58th Annual Meeting of the Association
  for Computational Linguistics}, pages 8440--8451.

\bibitem[{Conneau et~al.(2018)Conneau, Rinott, Lample, Williams, Bowman,
  Schwenk, and Stoyanov}]{conneau-etal-2018-xnli}
Alexis Conneau, Ruty Rinott, Guillaume Lample, Adina Williams, Samuel Bowman,
  Holger Schwenk, and Veselin Stoyanov. 2018.
\newblock {XNLI}: Evaluating cross-lingual sentence representations.
\newblock In \emph{Proceedings of the 2018 Conference on Empirical Methods in
  Natural Language Processing}, pages 2475--2485.

\bibitem[{Cooper et~al.(1994)Cooper, Crouch, van Eijck, Fox, van Genabith,
  Jaspers, Kamp, Pinkal, Poesio, Pulman et~al.}]{cooper1994fracas}
Robin Cooper, Richard Crouch, Jan van Eijck, Chris Fox, Josef van Genabith, Jan
  Jaspers, Hans Kamp, Manfred Pinkal, Massimo Poesio, Stephen Pulman, et~al.
  1994.
\newblock {F}ra{C}a{S}--a framework for computational semantics.
\newblock \emph{Deliverable}, D6.

\bibitem[{Dagan et~al.(2006)Dagan, Glickman, and Magnini}]{dagan2006}
Ido Dagan, Oren Glickman, and Bernardo Magnini. 2006.
\newblock The pascal recognising textual entailment challenge.
\newblock In \emph{Machine Learning Challenges. Evaluating Predictive
  Uncertainty, Visual Object Classification, and Recognising Tectual
  Entailment}, pages 177--190.

\bibitem[{Devlin et~al.(2019)Devlin, Chang, Lee, and
  Toutanova}]{devlin-etal-2019-bert}
Jacob Devlin, Ming-Wei Chang, Kenton Lee, and Kristina Toutanova. 2019.
\newblock {BERT}: Pre-training of deep bidirectional transformers for language
  understanding.
\newblock In \emph{Proceedings of the 2019 Conference of the North {A}merican
  Chapter of the Association for Computational Linguistics: Human Language
  Technologies, Volume 1 (Long and Short Papers)}, pages 4171--4186.

\bibitem[{Frege(1963)}]{frege1963compound}
Gottlob Frege. 1963.
\newblock Compound thoughts.
\newblock \emph{Mind}, 72(285):1--17.

\bibitem[{Gantt et~al.(2020)Gantt, Kane, and White}]{gantt-etal-2020-natural}
William Gantt, Benjamin Kane, and Aaron~Steven White. 2020.
\newblock Natural language inference with mixed effects.
\newblock In \emph{Proceedings of the Ninth Joint Conference on Lexical and
  Computational Semantics}, pages 81--87.

\bibitem[{Glockner et~al.(2018)Glockner, Shwartz, and
  Goldberg}]{glockner-etal-2018-breaking}
Max Glockner, Vered Shwartz, and Yoav Goldberg. 2018.
\newblock Breaking {NLI} systems with sentences that require simple lexical
  inferences.
\newblock In \emph{Proceedings of the 56th Annual Meeting of the Association
  for Computational Linguistics (Volume 2: Short Papers)}, pages 650--655.

\bibitem[{Goodwin et~al.(2020)Goodwin, Sinha, and
  O{'}Donnell}]{goodwin-etal-2020-probing}
Emily Goodwin, Koustuv Sinha, and Timothy~J. O{'}Donnell. 2020.
\newblock Probing linguistic systematicity.
\newblock In \emph{Proceedings of the 58th Annual Meeting of the Association
  for Computational Linguistics}, pages 1958--1969.

\bibitem[{Gupta et~al.(2021)Gupta, Kvernadze, and
  Srikumar}]{DBLP:conf/aaai/GuptaKS21}
Ashim Gupta, Giorgi Kvernadze, and Vivek Srikumar. 2021.
\newblock {BERT} {\&} family eat word salad: Experiments with text
  understanding.
\newblock In \emph{Thirty-Fifth {AAAI} Conference on Artificial Intelligence,
  {AAAI} 2021}, pages 12946--12954.

\bibitem[{Ham et~al.(2020)Ham, Choe, Park, Choi, and
  Soh}]{ham-etal-2020-kornli}
Jiyeon Ham, Yo~Joong Choe, Kyubyong Park, Ilji Choi, and Hyungjoon Soh. 2020.
\newblock {K}or{NLI} and {K}or{STS}: New benchmark datasets for {K}orean
  natural language understanding.
\newblock In \emph{Findings of the Association for Computational Linguistics:
  EMNLP 2020}, pages 422--430.

\bibitem[{Hawkins(1978)}]{Hawkins1978}
John Hawkins. 1978.
\newblock \emph{Definiteness and Indefiniteness. A Study in Reference and
  Grammaticality Prediction}.
\newblock Routledge.

\bibitem[{Hayashibe(2020)}]{hayashibe-2020-japanese}
Yuta Hayashibe. 2020.
\newblock {J}apanese realistic textual entailment corpus.
\newblock In \emph{Proceedings of the 12th Language Resources and Evaluation
  Conference}, pages 6827--6834.

\bibitem[{Heim(1982)}]{Heim1982}
Irene Heim. 1982.
\newblock \emph{The Semantics of Definite and Indefinite Noun Phrases}.
\newblock Ph.D. thesis, UMass Amherst.

\bibitem[{Hessel and Schofield(2021)}]{hessel-schofield-2021-effective}
Jack Hessel and Alexandra Schofield. 2021.
\newblock How effective is {BERT} without word ordering? implications for
  language understanding and data privacy.
\newblock In \emph{Proceedings of the 59th Annual Meeting of the Association
  for Computational Linguistics and the 11th International Joint Conference on
  Natural Language Processing (Volume 2: Short Papers)}, pages 204--211.

\bibitem[{Hinds(1986)}]{hinds1986}
John Hinds. 1986.
\newblock \emph{Japanese: Descriptive Grammar}.
\newblock Croom Helm.

\bibitem[{Hoji(1985)}]{hoji1985}
Hajime Hoji. 1985.
\newblock Logical form constraints and configurational structures in japanese.
\newblock \emph{PHD Thesis. University of Washington}.

\bibitem[{Hu et~al.(2020)Hu, Richardson, Xu, Li, K{\"u}bler, and
  Moss}]{hu-etal-2020-ocnli}
Hai Hu, Kyle Richardson, Liang Xu, Lu~Li, Sandra K{\"u}bler, and Lawrence Moss.
  2020.
\newblock {OCNLI}: {O}riginal {C}hinese {N}atural {L}anguage {I}nference.
\newblock In \emph{Findings of the Association for Computational Linguistics:
  EMNLP 2020}, pages 3512--3526.

\bibitem[{Janssen and Partee(1997)}]{janssen1997compositionality}
Theo Janssen and Barbara Partee. 1997.
\newblock Compositionality.
\newblock In Johan van Benthem and Alice ter Meulen, editors, \emph{Handbook of
  Logic and Language}, pages 417--473. Elsevier.

\bibitem[{Joshi et~al.(2020)Joshi, Santy, Budhiraja, Bali, and
  Choudhury}]{joshi-etal-2020-state}
Pratik Joshi, Sebastin Santy, Amar Budhiraja, Kalika Bali, and Monojit
  Choudhury. 2020.
\newblock The state and fate of linguistic diversity and inclusion in the {NLP}
  world.
\newblock In \emph{Proceedings of the 58th Annual Meeting of the Association
  for Computational Linguistics}, pages 6282--6293.

\bibitem[{Kalouli et~al.(2017)Kalouli, Real, and
  de~Paiva}]{kalouli-etal-2017-textual}
Aikaterini-Lida Kalouli, Livy Real, and Valeria de~Paiva. 2017.
\newblock Textual inference: getting logic from humans.
\newblock In \emph{Proceedings of the 12th International Conference on
  Computational Semantics (IWCS) {---} Short papers}.

\bibitem[{Katz and Fodor(1963)}]{katz1963structure}
Jerrold Katz and Jerry Fodor. 1963.
\newblock The structure of a semantic theory.
\newblock \emph{Language}, 39(2):170--210.

\bibitem[{Kawazoe et~al.(2017)Kawazoe, Tanaka, Mineshima, and
  Bekki}]{10.1007/978-3-319-50953-2_5}
Ai~Kawazoe, Ribeka Tanaka, Koji Mineshima, and Daisuke Bekki. 2017.
\newblock An inference problem set for evaluating semantic theories and
  semantic processing systems for japanese.
\newblock In \emph{New Frontiers in Artificial Intelligence}, pages 58--65.

\bibitem[{Kuribayashi et~al.(2021)Kuribayashi, Oseki, Ito, Yoshida, Asahara,
  and Inui}]{kuribayashi-etal-2021-lower}
Tatsuki Kuribayashi, Yohei Oseki, Takumi Ito, Ryo Yoshida, Masayuki Asahara,
  and Kentaro Inui. 2021.
\newblock Lower perplexity is not always human-like.
\newblock In \emph{Proceedings of the 59th Annual Meeting of the Association
  for Computational Linguistics and the 11th International Joint Conference on
  Natural Language Processing (Volume 1: Long Papers)}, pages 5203--5217.

\bibitem[{Le et~al.(2020)Le, Vial, Frej, Segonne, Coavoux, Lecouteux, Allauzen,
  Crabb{\'e}, Besacier, and Schwab}]{le-etal-2020-flaubert}
Hang Le, Lo{\"\i}c Vial, Jibril Frej, Vincent Segonne, Maximin Coavoux,
  Benjamin Lecouteux, Alexandre Allauzen, Benoit Crabb{\'e}, Laurent Besacier,
  and Didier Schwab. 2020.
\newblock {F}lau{BERT}: Unsupervised language model pre-training for {F}rench.
\newblock In \emph{Proceedings of the 12th Language Resources and Evaluation
  Conference}, pages 2479--2490.

\bibitem[{Levy and Andrew(2006)}]{levy-andrew-2006-tregex}
Roger Levy and Galen Andrew. 2006.
\newblock Tregex and tsurgeon: tools for querying and manipulating tree data
  structures.
\newblock In \emph{Proceedings of the Fifth International Conference on
  Language Resources and Evaluation ({LREC}{'}06)}.

\bibitem[{Liang et~al.(2020)Liang, Duan, Gong, Wu, Guo, Qi, Gong, Shou, Jiang,
  Cao, Fan, Zhang, Agrawal, Cui, Wei, Bharti, Qiao, Chen, Wu, Liu, Yang,
  Campos, Majumder, and Zhou}]{liang-etal-2020-xglue}
Yaobo Liang, Nan Duan, Yeyun Gong, Ning Wu, Fenfei Guo, Weizhen Qi, Ming Gong,
  Linjun Shou, Daxin Jiang, Guihong Cao, Xiaodong Fan, Ruofei Zhang, Rahul
  Agrawal, Edward Cui, Sining Wei, Taroon Bharti, Ying Qiao, Jiun-Hung Chen,
  Winnie Wu, Shuguang Liu, Fan Yang, Daniel Campos, Rangan Majumder, and Ming
  Zhou. 2020.
\newblock {XGLUE}: A new benchmark datasetfor cross-lingual pre-training,
  understanding and generation.
\newblock In \emph{Proceedings of the 2020 Conference on Empirical Methods in
  Natural Language Processing (EMNLP)}, pages 6008--6018.

\bibitem[{Linzen(2020)}]{linzen-2020-accelerate}
Tal Linzen. 2020.
\newblock How can we accelerate progress towards human-like linguistic
  generalization?
\newblock In \emph{Proceedings of the 58th Annual Meeting of the Association
  for Computational Linguistics}, pages 5210--5217.

\bibitem[{Liu et~al.(2019)Liu, Ott, Goyal, Du, Joshi, Chen, Levy, Lewis,
  Zettlemoyer, and Stoyanov}]{RoBERTa}
Yinhan Liu, Myle Ott, Naman Goyal, Jingfei Du, Mandar Joshi, Danqi Chen, Omer
  Levy, Mike Lewis, Luke Zettlemoyer, and Veselin Stoyanov. 2019.
\newblock Ro{BERT}a: A robustly optimized bert pretraining approach.
\newblock \emph{CoRR}, cs.CL/1907.11692.
\newblock Version 1.

\bibitem[{Marelli et~al.(2014)Marelli, Menini, Baroni, Bentivogli, Bernardi,
  and Zamparelli}]{marelli-etal-2014-sick}
Marco Marelli, Stefano Menini, Marco Baroni, Luisa Bentivogli, Raffaella
  Bernardi, and Roberto Zamparelli. 2014.
\newblock A {SICK} cure for the evaluation of compositional distributional
  semantic models.
\newblock In \emph{Proceedings of the Ninth International Conference on
  Language Resources and Evaluation ({LREC}'14)}, pages 216--223.

\bibitem[{McCoy et~al.(2019)McCoy, Pavlick, and Linzen}]{mccoy-etal-2019-right}
Tom McCoy, Ellie Pavlick, and Tal Linzen. 2019.
\newblock Right for the wrong reasons: Diagnosing syntactic heuristics in
  natural language inference.
\newblock In \emph{Proceedings of the 57th Annual Meeting of the Association
  for Computational Linguistics}, pages 3428--3448.

\bibitem[{Montague(1973)}]{Montague:1973}
Richard Montague. 1973.
\newblock The proper treatment of quantification in ordinary {English}.
\newblock In K.~J.~J. Hintikka, J.~Moravcsic, and P.~Suppes, editors,
  \emph{Approaches to Natural Language}, pages 221--242. Reidel, Dordrecht.

\bibitem[{Morita et~al.(2015)Morita, Kawahara, and
  Kurohashi}]{morita-etal-2015-morphological}
Hajime Morita, Daisuke Kawahara, and Sadao Kurohashi. 2015.
\newblock Morphological analysis for unsegmented languages using recurrent
  neural network language model.
\newblock In \emph{Proceedings of the 2015 Conference on Empirical Methods in
  Natural Language Processing}, pages 2292--2297.

\bibitem[{Naik et~al.(2018)Naik, Ravichander, Sadeh, Rose, and
  Neubig}]{naik-etal-2018-stress}
Aakanksha Naik, Abhilasha Ravichander, Norman Sadeh, Carolyn Rose, and Graham
  Neubig. 2018.
\newblock Stress test evaluation for natural language inference.
\newblock In \emph{Proceedings of the 27th International Conference on
  Computational Linguistics}, pages 2340--2353.

\bibitem[{Nakanishi and Tomioka(2004)}]{nakanishi2004japanese}
Kimiko Nakanishi and Satoshi Tomioka. 2004.
\newblock Japanese plurals are exceptional.
\newblock \emph{Journal of East Asian Linguistics}, 13(2):113--140.

\bibitem[{Park et~al.(2021)Park, Moon, Kim, Cho, Han, Park, Song, Kim, Song,
  Oh, Lee, Oh, Lyu, Jeong, Lee, Seo, Lee, Kim, Lee, Jang, Do, Kim, Lim, Lee,
  Park, Shin, Kim, Park, Oh, Ha, and Cho}]{park2021klue}
Sungjoon Park, Jihyung Moon, Sungdong Kim, Won~Ik Cho, Jiyoon Han, Jangwon
  Park, Chisung Song, Junseong Kim, Yongsook Song, Taehwan Oh, Joohong Lee,
  Juhyun Oh, Sungwon Lyu, Younghoon Jeong, Inkwon Lee, Sangwoo Seo, Dongjun
  Lee, Hyunwoo Kim, Myeonghwa Lee, Seongbo Jang, Seungwon Do, Sunkyoung Kim,
  Kyungtae Lim, Jongwon Lee, Kyumin Park, Jamin Shin, Seonghyun Kim, Lucy Park,
  Alice Oh, Jung-Woo Ha, and Kyunghyun Cho. 2021.
\newblock {KLUE}: {K}orean language understanding evaluation.
\newblock \emph{CoRR}, cs.CL/2105.09680.
\newblock Version 1.

\bibitem[{Pavlick and Kwiatkowski(2019)}]{10.1162/tacl_a_00293}
Ellie Pavlick and Tom Kwiatkowski. 2019.
\newblock {Inherent Disagreements in Human Textual Inferences}.
\newblock \emph{Transactions of the Association for Computational Linguistics},
  7:677--694.

\bibitem[{Pham et~al.(2021)Pham, Bui, Mai, and Nguyen}]{pham-etal-2021-order}
Thang Pham, Trung Bui, Long Mai, and Anh Nguyen. 2021.
\newblock Out of order: How important is the sequential order of words in a
  sentence in natural language understanding tasks?
\newblock In \emph{Findings of the Association for Computational Linguistics:
  ACL-IJCNLP 2021}, pages 1145--1160.

\bibitem[{Ravfogel et~al.(2019)Ravfogel, Goldberg, and
  Linzen}]{ravfogel-etal-2019-studying}
Shauli Ravfogel, Yoav Goldberg, and Tal Linzen. 2019.
\newblock Studying the inductive biases of {RNN}s with synthetic variations of
  natural languages.
\newblock In \emph{Proceedings of the 2019 Conference of the North {A}merican
  Chapter of the Association for Computational Linguistics: Human Language
  Technologies, Volume 1 (Long and Short Papers)}, pages 3532--3542.

\bibitem[{Real et~al.(2018)}]{real2018sickBR}
Livy Real et~al. 2018.
\newblock {SICK-BR}: A portuguese corpus for inference.
\newblock In \emph{Computational Processing of the Portuguese Language}, pages
  303--312.

\bibitem[{Rozen et~al.(2019)Rozen, Shwartz, Aharoni, and
  Dagan}]{rozen-etal-2019-diversify}
Ohad Rozen, Vered Shwartz, Roee Aharoni, and Ido Dagan. 2019.
\newblock Diversify your datasets: Analyzing generalization via controlled
  variance in adversarial datasets.
\newblock In \emph{Proceedings of the 23rd Conference on Computational Natural
  Language Learning (CoNLL)}, pages 196--205.

\bibitem[{Rust et~al.(2021)Rust, Pfeiffer, Vuli{\'c}, Ruder, and
  Gurevych}]{rust-etal-2021-good}
Phillip Rust, Jonas Pfeiffer, Ivan Vuli{\'c}, Sebastian Ruder, and Iryna
  Gurevych. 2021.
\newblock How good is your tokenizer? on the monolingual performance of
  multilingual language models.
\newblock In \emph{Proceedings of the 59th Annual Meeting of the Association
  for Computational Linguistics and the 11th International Joint Conference on
  Natural Language Processing (Volume 1: Long Papers)}, pages 3118--3135.

\bibitem[{Saito(1985)}]{saito1985some}
Mamoru Saito. 1985.
\newblock \emph{Some asymmetries in Japanese and their theoretical
  implications}.
\newblock Ph.D. thesis, NA Cambridge.

\bibitem[{Schuster and Nakajima(2012)}]{schuster2012japanese}
Mike Schuster and Kaisuke Nakajima. 2012.
\newblock Japanese and korean voice search.
\newblock In \emph{2012 IEEE International Conference on Acoustics, Speech and
  Signal Processing (ICASSP)}, pages 5149--5152.

\bibitem[{Seelawi et~al.(2021)Seelawi, Tuffaha, Gzawi, Farhan, Talafha, Badawi,
  Sober, Al-Dweik, Freihat, and Al-Natsheh}]{seelawi-etal-2021-alue}
Haitham Seelawi, Ibraheem Tuffaha, Mahmoud Gzawi, Wael Farhan, Bashar Talafha,
  Riham Badawi, Zyad Sober, Oday Al-Dweik, Abed~Alhakim Freihat, and Hussein
  Al-Natsheh. 2021.
\newblock {ALUE}: {A}rabic language understanding evaluation.
\newblock In \emph{Proceedings of the Sixth Arabic Natural Language Processing
  Workshop}, pages 173--184.

\bibitem[{Shavrina et~al.(2020)Shavrina, Fenogenova, Anton, Shevelev, Artemova,
  Malykh, Mikhailov, Tikhonova, Chertok, and
  Evlampiev}]{shavrina-etal-2020-russiansuperglue}
Tatiana Shavrina, Alena Fenogenova, Emelyanov Anton, Denis Shevelev, Ekaterina
  Artemova, Valentin Malykh, Vladislav Mikhailov, Maria Tikhonova, Andrey
  Chertok, and Andrey Evlampiev. 2020.
\newblock {R}ussian{S}uper{GLUE}: A {R}ussian language understanding evaluation
  benchmark.
\newblock In \emph{Proceedings of the 2020 Conference on Empirical Methods in
  Natural Language Processing (EMNLP)}, pages 4717--4726.

\bibitem[{Shibatani(1990)}]{shibatani1990languages}
Masayoshi Shibatani. 1990.
\newblock \emph{The Languages of {J}apan}.
\newblock Cambridge University Press.

\bibitem[{Sinha et~al.(2021{\natexlab{a}})Sinha, Jia, Hupkes, Pineau, Williams,
  and Kiela}]{sinha2021masked}
Koustuv Sinha, Robin Jia, Dieuwke Hupkes, Joelle Pineau, Adina Williams, and
  Douwe Kiela. 2021{\natexlab{a}}.
\newblock Masked language modeling and the distributional hypothesis: Order
  word matters pre-training for little.
\newblock \emph{CoRR}, cs.CL/2104.06644.
\newblock Version 1.

\bibitem[{Sinha et~al.(2021{\natexlab{b}})Sinha, Parthasarathi, Pineau, and
  Williams}]{sinha2021unnat}
Koustuv Sinha, Prasanna Parthasarathi, Joelle Pineau, and Adina Williams.
  2021{\natexlab{b}}.
\newblock {UnNatural Language Inference}.
\newblock In \emph{Proceedings of the Joint Conference of the 59th Annual
  Meeting of the Association for Computational Linguistics and the 11th
  International Joint Conference on Natural Language Processing
  (ACL-IJCNLP2021)}.

\bibitem[{Tolmachev et~al.(2018)Tolmachev, Kawahara, and
  Kurohashi}]{tolmachev-etal-2018-juman}
Arseny Tolmachev, Daisuke Kawahara, and Sadao Kurohashi. 2018.
\newblock {J}uman++: A morphological analysis toolkit for scriptio continua.
\newblock In \emph{Proceedings of the 2018 Conference on Empirical Methods in
  Natural Language Processing: System Demonstrations}, pages 54--59.

\bibitem[{Wang et~al.(2019)Wang, Singh, Michael, Hill, Levy, and
  Bowman}]{wang2018glue}
Alex Wang, Amanpreet Singh, Julian Michael, Felix Hill, Omer Levy, and Samuel
  Bowman. 2019.
\newblock {GLUE}: A multi-task benchmark and analysis platform for natural
  language understanding.
\newblock In \emph{Proceedings of the International Conference on Learning
  Representations (ICLR)}.

\bibitem[{White and Cotterell(2021)}]{white-cotterell-2021-examining}
Jennifer~C. White and Ryan Cotterell. 2021.
\newblock Examining the inductive bias of neural language models with
  artificial languages.
\newblock In \emph{Proceedings of the 59th Annual Meeting of the Association
  for Computational Linguistics and the 11th International Joint Conference on
  Natural Language Processing (Volume 1: Long Papers)}, pages 454--463.

\bibitem[{Wijnholds and Moortgat(2021)}]{wijnholds2021sicknl}
Gijs Wijnholds and Michael Moortgat. 2021.
\newblock {SICK}-{NL}: A dataset for {D}utch natural language inference.
\newblock In \emph{Proceedings of the 16th Conference of the European Chapter
  of the Association for Computational Linguistics: Main Volume}, pages
  1474--1479.

\bibitem[{Williams et~al.(2018)Williams, Nangia, and
  Bowman}]{williams-etal-2018-broad}
Adina Williams, Nikita Nangia, and Samuel Bowman. 2018.
\newblock A broad-coverage challenge corpus for sentence understanding through
  inference.
\newblock In \emph{Proceedings of the 2018 Conference of the North {A}merican
  Chapter of the Association for Computational Linguistics: Human Language
  Technologies, Volume 1 (Long Papers)}, pages 1112--1122.

\bibitem[{Xu et~al.(2020)Xu, Hu, Zhang, Li, Cao, Li, Xu, Sun, Yu, Yu, Tian,
  Dong, Liu, Shi, Cui, Li, Zeng, Wang, Xie, Li, Patterson, Tian, Zhang, Zhou,
  Liu, Zhao, Zhao, Yue, Zhang, Yang, Richardson, and Lan}]{xu-etal-2020-clue}
Liang Xu, Hai Hu, Xuanwei Zhang, Lu~Li, Chenjie Cao, Yudong Li, Yechen Xu, Kai
  Sun, Dian Yu, Cong Yu, Yin Tian, Qianqian Dong, Weitang Liu, Bo~Shi, Yiming
  Cui, Junyi Li, Jun Zeng, Rongzhao Wang, Weijian Xie, Yanting Li, Yina
  Patterson, Zuoyu Tian, Yiwen Zhang, He~Zhou, Shaoweihua Liu, Zhe Zhao, Qipeng
  Zhao, Cong Yue, Xinrui Zhang, Zhengliang Yang, Kyle Richardson, and Zhenzhong
  Lan. 2020.
\newblock {CLUE}: A {C}hinese language understanding evaluation benchmark.
\newblock In \emph{Proceedings of the 28th International Conference on
  Computational Linguistics}, pages 4762--4772.

\bibitem[{Yanaka et~al.(2021)Yanaka, Mineshima, and
  Inui}]{yanaka-etal-2021-exploring}
Hitomi Yanaka, Koji Mineshima, and Kentaro Inui. 2021.
\newblock Exploring transitivity in neural {NLI} models through veridicality.
\newblock In \emph{Proceedings of the 16th Conference of the European Chapter
  of the Association for Computational Linguistics: Main Volume}, pages
  920--934.

\bibitem[{Yang et~al.(2019)Yang, Zhang, Tar, and
  Baldridge}]{yang-etal-2019-paws}
Yinfei Yang, Yuan Zhang, Chris Tar, and Jason Baldridge. 2019.
\newblock {PAWS}-{X}: A cross-lingual adversarial dataset for paraphrase
  identification.
\newblock In \emph{Proceedings of the 2019 Conference on Empirical Methods in
  Natural Language Processing and the 9th International Joint Conference on
  Natural Language Processing (EMNLP-IJCNLP)}, pages 3687--3692.

\bibitem[{Yoshikawa et~al.(2017)Yoshikawa, Noji, and
  Matsumoto}]{yoshikawa:2017acl}
Masashi Yoshikawa, Hiroshi Noji, and Yuji Matsumoto. 2017.
\newblock {A}* {CCG} parsing with a supertag and dependency factored model.
\newblock In \emph{Proceedings of the 55th Annual Meeting of the Association
  for Computational Linguistics (Volume 1: Long Papers)}, pages 277--287.

\bibitem[{Yoshikoshi et~al.(2020)Yoshikoshi, Kawahara, and Kurohashi}]{jsnli}
Takumi Yoshikoshi, Daisuke Kawahara, and Sadao Kurohashi. 2020.
\newblock Multilingualization of natural language inference datasets using
  machine translation (in {J}apanese).
\newblock In \emph{The 244th Meeting of Natural Language Processing}.

\bibitem[{Zhang et~al.(2020)Zhang, Kishore, Wu, Weinberger, and
  Artzi}]{bert-score}
Tianyi Zhang, Varsha Kishore, Felix Wu, Kilian~Q. Weinberger, and Yoav Artzi.
  2020.
\newblock Bertscore: Evaluating text generation with bert.
\newblock In \emph{Proceedings of the International Conference on Learning
  Representations (ICLR)}.

\bibitem[{Zhang et~al.(2019)Zhang, Baldridge, and He}]{zhang-etal-2019-paws}
Yuan Zhang, Jason Baldridge, and Luheng He. 2019.
\newblock {PAWS}: Paraphrase adversaries from word scrambling.
\newblock In \emph{Proceedings of the 2019 Conference of the North {A}merican
  Chapter of the Association for Computational Linguistics: Human Language
  Technologies, Volume 1 (Long and Short Papers)}, pages 1298--1308.

\end{thebibliography}
\bibliographystyle{acl_natbib}

\clearpage
\appendix

\end{document}